\documentclass{article}

\usepackage{arxiv}

\usepackage[utf8]{inputenc} % allow utf-8 input
\usepackage[T1]{fontenc}    % use 8-bit T1 fonts
\usepackage{hyperref}       % hyperlinks
\usepackage{url}            % simple URL typesetting
\usepackage{booktabs}       % professional-quality tables
\usepackage{amsfonts}       % blackboard math symbols
\usepackage{nicefrac}       % compact symbols for 1/2, etc.
\usepackage{microtype}      % microtypography
\usepackage{lipsum}
\usepackage{listings}
\usepackage{graphicx}
\graphicspath{ {./images/} }
\usepackage{amssymb}
\usepackage{amsmath}
\usepackage{natbib}
\usepackage{algorithm}
\usepackage{algpseudocode}
\usepackage{multirow}
\usepackage{caption}
\usepackage{tabularx,array,listings,xcolor,float}
\definecolor{sagehostbg}{RGB}{240,247,252}
\definecolor{sagetilingbg}{RGB}{251,247,235}
\definecolor{sagekernelbg}{RGB}{240,248,242}
\definecolor{sageinstructionbg}{RGB}{246,246,246}

\lstdefinestyle{sagecpp}{
  language=C++,
  basicstyle=\ttfamily\scriptsize,
  keywordstyle=\color{blue!65!black},
  commentstyle=\color{green!35!black},
  stringstyle=\color{orange!55!black},
  numbers=left,
  numberstyle=\tiny\color{black!45},
  numbersep=6pt,
  frame=single,
  framerule=0.25pt,
  rulecolor=\color{black!20},
  breaklines=true,
  breakatwhitespace=false,
  columns=fullflexible,
  keepspaces=true,
  showstringspaces=false,
  tabsize=2,
  xleftmargin=1.5em,
  framexleftmargin=1.2em,
  aboveskip=0.6em,
  belowskip=0.8em
}

\lstdefinestyle{sageinstruction}{
  basicstyle=\ttfamily\small,
  backgroundcolor=\color{sageinstructionbg},
  frame=single,
  framerule=0.25pt,
  rulecolor=\color{black!20},
  breaklines=true,
  breakatwhitespace=false,
  breakindent=0pt,
  breakautoindent=false,
  columns=fullflexible,
  keepspaces=true,
  showstringspaces=false,
  xleftmargin=0pt,
  xrightmargin=0pt,
  framexleftmargin=0pt,
  framexrightmargin=0pt,
  framesep=3pt,
  aboveskip=0.5em,
  belowskip=0.8em
}

\title{From Experience to Expertise: {Adoption-Aware} Memory Learning for Data-Scarce NPU Kernel Synthesis}

\author{
 Longxiao Fan \\
  Fudan University \\
  \texttt{24110240020@m.fudan.edu.cn} \\
   \And
 Tao Zhang \\
  University of Science and Technology of China \\
  \texttt{zhangtaolqy@mail.ustc.edu.cn} \\
  \And
 Han Yan \\
  Fudan University \\
  \texttt{24210240065@m.fudan.edu.cn} \\
  \And
 Jiajun Li \\
  Huawei Technologies Ltd. \\
  \texttt{jiajun.work@huawei.com} \\
  \And
 Mingcong Song \\
  Huawei Technologies Ltd. \\
  \texttt{songmingcong@huawei.com} \\
  \And
 Guoping Long \\
  Huawei Technologies Ltd. \\
  \texttt{robin3@huawei.com} \\
  \And
 Hongjie Si \\
  Huawei Technologies Ltd. \\
  \texttt{sihongjie@huawei.com} \\
  \And
 Weiwei Sun \\
  Fudan University \\
  \texttt{wwsun@fudan.edu.cn} \\
}

\newcolumntype{Y}{>{\raggedright\arraybackslash}X}
\newcommand{\1}{\mathbf{1}}
\begin{document}
\maketitle
\begin{abstract}
High-performance kernels underpin efficient accelerator execution but require expert tuning and lengthy manual optimization cycles. LLM coding agents promise automation, yet their CUDA knowledge transfers poorly to data-scarce domain-specific architectures (DSAs) such as NPUs, whose execution models and memory hierarchies differ substantially from those of GPUs. To address this transfer gap, post-training methods adapt LLMs to NPU programming but depend on scarce expert data and substantial training compute. Memory-learning agents instead adapt through external memory, but their uniform credit assignment gives adopted and unused experiences the same reward target, potentially biasing subsequent retrieval rankings. Moreover, when learned values guide only retrieval, high-value experiences that generalize across operators must be retrieved repeatedly rather than retained in context, thereby increasing retrieval overhead and weakening cross-task guidance. We therefore present SAGE, a persistent self-improving agent for NPU kernel synthesis. Adoption-Traced Utility estimation (ATU) combines explicit adoption records with kernel evaluation outcomes for adoption-aware credit assignment. Utility-Gated Consolidation (UGC) uses positive utility and repeated adoption across operators to select and abstract reusable rules into a bounded resident context. On NPUKernelBench, SAGE achieves a 95.5\% execution rate versus 84.1\% for the strongest controlled baseline, with 86.9\% of solved operators outperforming torch\_npu. With GLM-5.3, SAGE achieves a $43.99\times$ speedup over the torch\_npu reference on sparse flash attention. These results show that adoption-aware credit assignment and selective consolidation enable agents to accumulate and reuse hardware-specific knowledge across tasks.

\end{abstract}

% keywords can be removed

\section{Introduction}
Large Language Models (LLMs) have made rapid progress in CUDA and Triton
kernel synthesis \citep{yu2026towards,sun2026cuda,dai2026cuda}, benefiting from mature ecosystems with abundant public kernels, documentation, and optimization idioms \citep{qiu2026agenticcann}. Such resources serve as both large-scale pre-training signals and readily available supervision or retrieval sources for post-training and augmentation, allowing LLMs to achieve strong zero-shot performance and be further specialized with relatively low cost \citep{arakelyan2023exploring}. In contrast, emerging domain-specific architectures (DSAs) such as Ascend NPUs \citep{silvano2025survey,liao2021ascend} provide far less public code and require architecture-specific reasoning over memory hierarchies and execution models \citep{zhou2025squeezing}. This distribution
shift is substantial: GPT-5.2 drops from 92\% CUDA correctness to 14\% Ascend C correctness on KernelBench L1 tasks \citep{ouyang2025kernelbench,zheng2026towards}.
Thus, data-scarce kernel synthesis depends not only on generation but also on acquiring and reusing hardware-specific expertise under scarce supervision.

\begin{figure}[H]
    \centering
    \includegraphics[width=\linewidth]{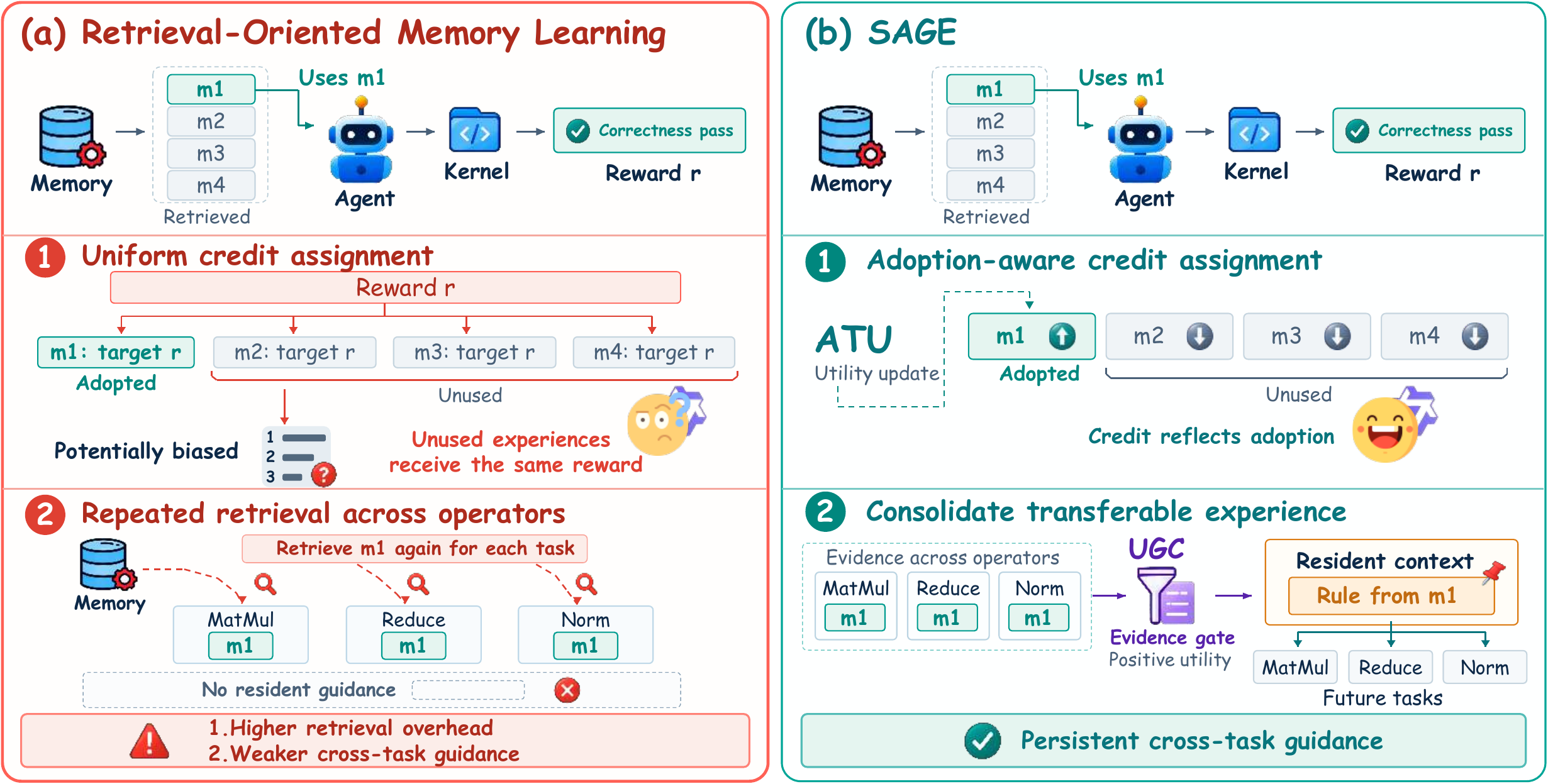}
   \caption{Retrieval-oriented memory learning with uniform credit and repeated retrieval (left) versus SAGE's adoption-aware credit assignment and selective consolidation into resident context (right).}
    \label{fig:motivation}
\end{figure}

To address scarce NPU programming knowledge, post-training methods adapt LLMs through supervised fine-tuning (SFT) \citep{zhou2023lima,chung2024scaling} on hardware-aware data and reinforcement learning (RL) \citep{zhang2025settling} with execution-based preferences \citep{cao2026ascendkernelgen}. Their scalability, however, is limited by scarce expert data and substantial training compute. Knowledge-augmented agents instead keep the backbone frozen and guide kernel generation and refinement with mined optimization experience \citep{wu2026ascendoptimizer} or structured domain knowledge \citep{qiu2026agenticcann}. Applying this knowledge to diagnose failures and revise kernels also generates new experience, raising the question of which experiences should guide subsequent tasks. 
Memory-learning methods in the broader agent literature address this through outcome-based values or dependency-aware credit propagation \citep{zhou2025memento,zhang2026memrl,liao2026memq}. EvoKernel applies memory learning to NPU kernel synthesis but uses \textbf{uniform credit assignment}, a coarse-grained rule that gives every retrieved experience the same verifier reward target within an iteration \citep{zheng2026towards}. A successful code repair can thus reward all top-$k$ retrieved experiences even when the agent adopts only a small subset. Such uniform reward targets can give unused experiences favorable value estimates and higher priority in later retrieval, potentially crowding out more promising experiences with lower scores. Moreover, when learned values guide only retrieval, high-value experiences that generalize across operators must be retrieved again rather than retained in context, potentially adding retrieval overhead and limiting consistent guidance across tasks. Figure~\ref{fig:motivation} contrasts \emph{retrieval-oriented memory learning} with SAGE's finer-grained \textbf{adoption-aware credit assignment} and selective consolidation into a bounded resident context.

Our central insight is to ground credit assignment and knowledge consolidation in evidence of use. Retrieval makes an experience available; recorded adoption indicates that it guided diagnosis or code revision. Adoption-aware credit assignment combines this evidence with the kernel outcome so that unused experiences do not receive the same credit as those adopted merely for being retrieved. A single adoption, however, does not establish transferability. Positive utility accumulated through repeated adoption across distinct operators provides evidence that an experience may encode a reusable NPU programming rule. Consolidating such rules into a bounded resident context makes them available to later tasks without requiring another successful retrieval.

Building on this insight, we propose \textbf{SAGE}, a self-improvement agent for NPU kernel generation and evolution. SAGE enables a frozen backbone LLM to evolve external memory through two complementary mechanisms. Adoption-Traced Utility estimation (ATU) combines explicit adoption records with episode-level verifier outcomes to update experience utility through adoption-aware credit assignment. Utility-Gated Consolidation (UGC) leverages positive utility and repeated adoption across operators to identify experiences for consolidation into reusable resident rules, while demoting underused rules under a fixed context budget. 
% SAGE further 
% incorporates cold-start knowledge construction and device-level diagnostic feedback enrichment to provide initial domain grounding and more informative execution signals during iterative refinement. 

Our contributions are as follows:

\begin{itemize}
    \item We introduce \textbf{SAGE}, a self-improving agent for NPU kernel synthesis that enables frozen LLMs to accumulate and reorganize hardware-specific knowledge across tasks.
    \item We propose \textbf{Adoption-Traced Utility estimation (ATU)}, an experience-level credit assignment mechanism that derives empirical utility estimates from adoption records.
    \item We propose \textbf{Utility-Gated Consolidation (UGC)}, a utility-driven memory evolution mechanism that transforms reusable experiences into persistent knowledge.
    \item We evaluate SAGE across three backbones on NPUKernelBench, demonstrating consistent gains in kernel correctness and runtime performance.
\end{itemize}

\section{Related Work}
\label{app:related_work}
\subsection{LLM-based Automated Kernel Synthesis}
LLM-based kernel synthesis has advanced rapidly in mature GPU ecosystems. CUDA-L1 \citep{sun2026cuda}, CudaForge \citep{zhang2025cudaforge}, KernelEvolve \citep{liao2026kernelevolve}, CUDA-Agent \citep{dai2026cuda}, Dr.Kernel \citep{liu2026dr}, and QiMeng-Kernel \citep{zhu2026qimeng} show that LLMs can generate and iteratively optimize CUDA or Triton kernels using compiler feedback, multi-agent coordination, and hardware-in-the-loop evaluation. Their success, however, benefits from abundant public kernels, mature toolchains, and well-documented optimization idioms. For data-scarce accelerators, AscendKernelGen \citep{cao2026ascendkernelgen} and NKI-Agent \citep{tang2026nki} use SFT and RL to internalize Ascend C or NKI programming knowledge, but their scalability remains constrained by expert data and training compute. AscendCraft \citep{wen2026ascendcraft} instead guides generation through a lightweight DSL and structured transcompilation into Ascend C. EvoKernel \citep{zheng2026towards}, AscendOptimizer \citep{wu2026ascendoptimizer}, AgenticCANN \citep{qiu2026agenticcann}, Hawk \citep{wen2026hawk} and AccelOpt \citep{zhang2026accelopt} retain frozen backbones and rely on retrieved knowledge, execution feedback, and iterative repair. These approaches reduce reliance on zero-shot generation, but lack explicit adoption records to guide credit assignment and the selection of reusable knowledge for persistent context.

\subsection{Memory-Augmented Agentic Self-Improvement}
LLM-based agents increasingly adapt through external memory without modifying backbone parameters. Reflexion \citep{shinn2023reflexion} converts task feedback into textual reflections retained across trials, while Voyager \citep{wang2023voyager} accumulates successful behaviors in a reusable skill library. ExpeL \citep{zhao2024expel} further distills transferable insights from collections of successful and failed trajectories. More recent work learns memory selection from downstream outcomes: Memento \citep{zhou2025memento} trains a reward-conditioned case-selection policy, while MemRL \citep{zhang2026memrl} learns reward-conditioned value estimates over retrieved experiences to guide subsequent selection. MemQ \citep{liao2026memq} further propagates outcome-based credit through provenance DAGs of memory-construction dependencies. Together, these methods improve experience selection and credit propagation, but their value estimates and provenance records do not reveal which experiences within a retrieved set actually informed a code revision. SAGE instead traces adoption to assign credit to individual experiences, then uses accumulated cross-task evidence to consolidate
repeatedly validated knowledge into persistent memory.

\section{Methods}
\subsection{Preliminaries}
We formulate NPU kernel synthesis over an ordered task stream under
\emph{non-parametric continual learning} \citep{gutierrez2025rag}, where a fixed generator repeatedly solves tasks while adapting only its persistent external memory.
For a kernel synthesis task $\tau=(x,g)$, $x$ denotes the operator specification and $g\in\{\mathrm{corr},\mathrm{opt}\}$ denotes the task-specific objective, corresponding respectively to a correctness episode and an optimization episode, which together form its synthesis trajectory.

Let $\mathcal{G}_{\theta_0}$ denote an LLM-based kernel generator whose parameters $\theta_0$ remain fixed throughout the task stream, and let $\mathcal{E}$ denote the compiler-and-hardware environment. At refinement round $t$, a generic context-construction procedure $\Gamma$ derives a memory-conditioned context: 
\begin{equation}
c_t=\Gamma\!\left(\mathcal{M},\tau,k_{t-1},f_{t-1}\right).
\end{equation}
Here, $\mathcal{M}$ is the persistent external memory state available before processing task $\tau$, while $k_{t-1}$ and $f_{t-1}$ 
denote the kernel implementation entering round $t$ and its latest evaluation feedback, respectively. At $t=1$, $(k_{0},f_{0})$ denotes an initial kernel--feedback pair and defaults to $(\varnothing,\varnothing)$ when unavailable. 
Conditioned on these inputs, the generator samples a candidate kernel and receives execution feedback evaluated by $\mathcal{E}$:
\begin{equation}
k_t = \mathcal{G}_{\theta_0} \!\left(\tau,k_{t-1},f_{t-1},c_t \right),\quad f_t= \mathcal{E}(\tau,k_t).
\end{equation}
The episode $\varepsilon=((c_t,k_t,f_t))_{t=1}^{T}$ terminates when the
objective specified by $g$ is met or the interaction budget is exhausted, where $T$ is the realized number of refinement rounds.
After the episode, the external memory state is updated as
\begin{equation}
\mathcal{M}\leftarrow\Phi(\mathcal{M},\varepsilon),
\end{equation}
where $\Phi$ denotes a generic non-parametric memory-update procedure. Repeated application of this transition across successive tasks constitutes the continual learning process. The objective is to improve subsequent execution outcomes solely via these memory transitions while $\theta_0$ remains fixed.

\subsection{SAGE Framework}
\label{sec:sage_framework}
\begin{figure}[H]
    \centering
    \includegraphics[width=\textwidth]{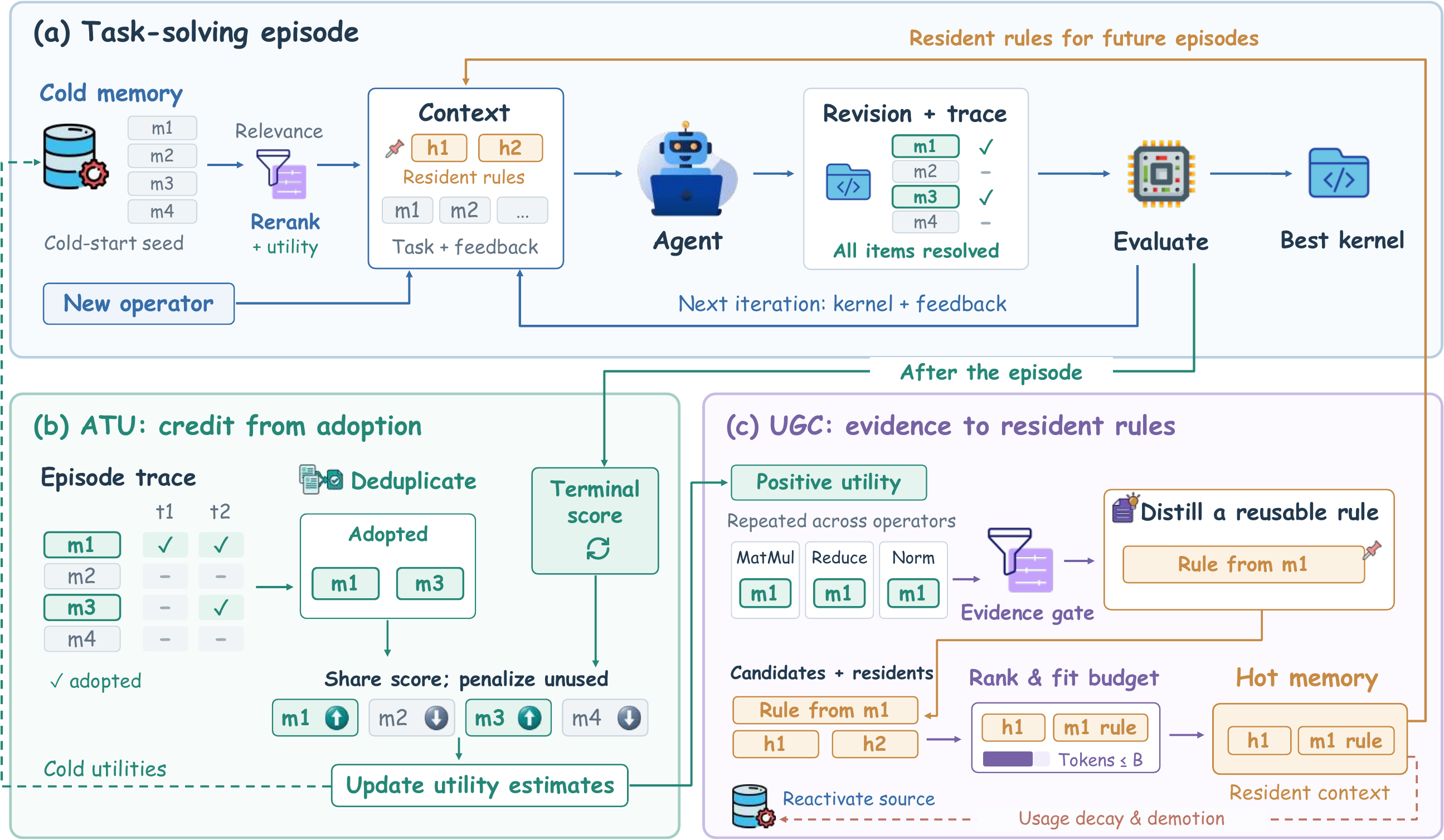}
    \caption{\textbf{Overview of SAGE.}
\textbf{(a)} A frozen agent iteratively refines kernels using retrieved cold experiences, resident hot rules, and execution feedback.
\textbf{(b)} ATU combines adoption traces with terminal outcomes to estimate item-level utility.
\textbf{(c)} UGC uses utility and repeated adoption across operators to select reusable rules for a bounded resident context and demote underused rules.}
    \label{fig:sage_framework}
\end{figure}

As shown in Figure~\ref{fig:sage_framework}, SAGE draws inspiration from
hierarchical memory management for fixed-context LLM agents
\citep{packer2023memgpt} and instantiates the generic memory state
$\mathcal{M}$ as a cold--hot hierarchy with associated evidence statistics:
\begin{equation}
    \mathcal{M} =
    \left(\mathcal{C},\mathcal{H}\right),
    \qquad
    \operatorname{Tok}(\mathcal{H}) \leq B,
\end{equation}
where $\mathcal{C}$ denotes \emph{cold memory}, which preserves source experiences and their item-level evidence and lifecycle statistics, while $\mathcal{H}$ denotes \emph{hot memory}, which provides consolidated rules to the generator's context without retrieval. Here, $B$ is the fixed token budget. At the end of each task, SAGE writes newly acquired experiences to cold memory $\mathcal{C}$ for retrieval in subsequent tasks.

SAGE realizes the memory transition $\Phi$ through two complementary mechanisms:
\begin{itemize}
    \item \textbf{Adoption-Traced Utility Estimation (ATU).}
    ATU uses adoption records to distribute terminal verifier credit among adopted experiences and penalize unused retrievals, thereby updating empirical utility estimates.
    \item \textbf{Utility-Gated Consolidation (UGC).}
    UGC uses positive utility and repeated adoption across operators to identify experiences for rule abstraction. It selects resident rules under a fixed token budget, allowing persistently underused rules to be displaced.
\end{itemize}

\subsection{Adoption-Traced Utility Estimation}
\label{sec:atu}
Compiler and device feedback is available after every generation, but it
evaluates the resulting kernel rather than the textual experiences that shaped
it. Uniform credit assignment applies the same reward target to all retrieved experiences, although only a subset may guide diagnosis or code revision. ATU performs adoption-aware credit assignment by combining explicit adoption records with terminal verifier outcomes to distinguish adopted experiences from unused retrievals.

\noindent\textbf{Trace-Grounded Experience Collection.}
For task $\tau$, let $\mathcal{R}_t\subseteq\mathcal{C}$ denote the experiences retrieved at refinement round $t$. For each item $m\in\mathcal{R}_t$, the agent records an adoption state $a_{m,t}\in\{1,0,\bot\}$ (adopted, not adopted, or unresolved) and a rationale.
To ensure complete adoption records, SAGE inserts a \emph{trace-completeness hook} in the evaluation script that blocks evaluation until the adoption state of every $m$ is explicitly resolved. Appendix~\ref{app:adoption_audit} reports an expert audit assessing whether recorded adoption corresponds to actual code revisions.
For each item, SAGE maintains deduplicated retrieval and adoption counts, $n_m^{\mathrm{ret}}$ and $n_m^{\mathrm{ado}}$, and the set $\mathcal{O}_m$ of distinct adopting operators.

\noindent\textbf{Adoption-Aware Credit Assignment.}
The benefit of an adopted experience may emerge only in the final kernel;
ATU therefore defers credit assignment until episode termination.
For an optimization episode, let $T_{\mathrm{first}}$ and $T_{\mathrm{best}}$
denote the execution latencies of the first correct kernel and the best correct
kernel found within the episode, respectively. We define the terminal score as
\begin{equation}
z =
\begin{cases}
1, & g=\mathrm{corr}\ \text{and correctness passes},\\[2pt]
\displaystyle
1-\frac{T_{\mathrm{best}}}{T_{\mathrm{first}}},
& g=\mathrm{opt}\ \text{and}\ T_{\mathrm{best}}<T_{\mathrm{first}},\\[7pt]
-\delta, & \text{otherwise},
\end{cases}
\label{eq:episode_score}
\end{equation}
where $0<\delta<1$ is a small failure penalty.
The optimization score measures the relative latency reduction achieved after
correctness is first established and is naturally bounded in $(0,1)$.
Let $\mathcal{R}=\bigcup_{t=1}^T\mathcal{R}_t$ and $\mathcal{A}=\{m\in\mathcal{R}:\exists t,\ a_{m,t}=1\}$ denote the retrieved and adopted experiences over episode $\varepsilon$. ATU converts the terminal score into item-level evidence:
\begin{equation}
\xi_{m} =
\begin{cases}
\displaystyle \frac{z}{|\mathcal{A}|},
    & m\in\mathcal{A}, \\[6pt]
-\beta,
    & m\in\mathcal{R}\setminus\mathcal{A},
\end{cases}
\end{equation}
where $0<\beta<1$ controls the penalty for retrieved but unused experiences. Hence, adopted items share a fixed episode-level credit budget, preventing credit inflation when multiple experiences are adopted, whereas non-adopted retrievals receive a fixed negative credit.

\noindent\textbf{Online Utility Tracking and Retrieval.}
Each memory item maintains an empirical utility $u_m\in[-1,1]$, initialized at zero and updated online via the following \emph{bandit-style} rule:
\begin{equation}
u_m \leftarrow
(1-\eta_m)u_m+\eta_m\xi_{m},
\qquad
\eta_m=\max\!\left(\frac{1}{1+n_m^{\mathrm{ret}}},\eta_{\min}\right).
\end{equation}
The decaying step size stabilizes utility estimates as evidence accumulates, while $0<\eta_{\min}\ll1$ prevents historical evidence from permanently dominating subsequent utility updates. Thus, $u_m$ estimates retrieval-conditioned empirical utility and subsequently guides experience ranking and memory consolidation.

At refinement round $t$, SAGE first uses a fixed hybrid sparse--dense retriever to obtain an over-retrieved candidate pool $\widetilde{\mathcal{R}}_t$ from cold memory based solely on relevance.
Utility is incorporated only during late reranking:
\begin{equation}
s(m)=s_{\mathrm{rel}}(m)\bigl(1+\gamma u_m\bigr),
\qquad m\in\widetilde{\mathcal{R}}_t,
\end{equation}
where $s_{\mathrm{rel}}(m)$ is the relevance score and $0\le\gamma<1$ controls the strength of utility modulation. The highest-ranked experiences form the final retrieval set $\mathcal{R}_t$ used for context construction and subsequent adoption tracing, while resident knowledge in $\mathcal{H}$ bypasses retrieval.

\subsection{Utility-Gated Consolidation}
\label{sec:ugc}

High-value experiences that generalize across operators should remain available in future task contexts without repeated retrieval. Utility-Gated Consolidation (UGC) uses positive utility and repeated adoption across operators to identify candidates for rule abstraction, then selects resident rules under a fixed context budget and allows underused rules to be displaced. UGC is invoked after each task-solving episode, following the ATU utility update.

\noindent\textbf{Evidence-Gated Admission and Abstraction.}
A cold-memory item becomes eligible for consolidation only after positive utility, sufficient adoption and cross-operator validation:
\begin{equation}
u_m>0,
\quad
n_m^{\mathrm{ado}}\ge K_{\mathrm{ado}},
\quad
|\mathcal{O}_m|\ge K_{\mathrm{op}},
\quad
K_{\mathrm{ado}},K_{\mathrm{op}}>1.
\end{equation}
Each source experience maintains a consolidation state
$\sigma_m \in
\{\mathrm{normal},\mathrm{validated},\mathrm{consolidated}\}$.
An item satisfying the above gates transitions from
$\mathrm{normal}$ to $\mathrm{validated}$, and the agent abstracts it into a concise transferable rule, optionally with a minimal code or API example, while preserving its source identifier. The source experience and its accumulated evidence are retained for provenance, but the source is excluded from active retrieval while its consolidated rule remains resident.
Source and rule share one identifier and utility $u_m$; the rule has no separate value estimator.

\noindent\textbf{Budgeted Hot-Memory Selection.}
Let $\mathcal{V}\subseteq\mathcal{C}$ denote the validated cold-memory
candidates eligible for consolidation. UGC ranks items in
$\mathcal{H}\cup\mathcal{V}$ by expected utility per resident token:
\begin{equation}
U(m)=\frac{u_m\,\hat p_m}{L_m},
\quad
\hat p_m=\frac{n_m^{\mathrm{ret}}}{N_m^{\mathrm{elig}}},
\quad m\in\mathcal{H}\cup\mathcal{V},
\end{equation}
where $\hat p_m$ is an empirical access-frequency estimate, $L_m$ is the resident token cost, and $N_m^{\mathrm{elig}}$ counts the episodes in which $m$ is active in cold memory and available to the retriever. Once item $m$ is written to cold memory, $N_m^{\mathrm{elig}}$ increases once per such episode, while $n_m^{\mathrm{ret}}$ increases only when $m$ is retrieved. The hot-memory set is greedily selected in descending $U(m)$ subject to $\operatorname{Tok}(\mathcal{H}) \le B$, where $B$ is the fixed token budget. The complete replacement procedure is given in Algorithm~\ref{alg:ugc}.
% \begin{algorithm}[t]
% \caption{Budgeted Hot-Memory Replacement}
% \label{alg:ugc}
% \begin{algorithmic}[1]
% \Require Current hot memory $\mathcal{H}$, validated candidates
% $\mathcal{V}$, utility density $U(\cdot)$, token budget $B$
% \Ensure Updated hot memory $\mathcal{H}'$

% \State $\mathcal{P} \gets \mathcal{H} \cup \mathcal{V}$
% \State $\mathcal{H}' \gets \varnothing$
% \ForAll{$m \in \mathcal{P}$ sorted by decreasing $U(m)$}
%     \If{$\operatorname{Tok}(\mathcal{H}') + L_m \le B$}
%         \State $\mathcal{H}' \gets \mathcal{H}' \cup \{m\}$
%     \EndIf
% \EndFor

% \State $\sigma_m \gets \mathrm{consolidated},
% \ \forall m\in\mathcal{H}'\setminus\mathcal{H}$
% \State $\sigma_m \gets \mathrm{validated},
% \ \forall m\in\mathcal{H}\setminus\mathcal{H}'$

% \State \Return $\mathcal{H}'$
% \end{algorithmic}
% \end{algorithm}

Resident items bypass retrieval, so the \emph{trace-completeness hook} introduced in Sec.~\ref{sec:atu} additionally records their episode-level usage. Their access-frequency estimates are updated by usage as
\begin{equation}
\hat p_m\leftarrow
(1-\mu)\hat p_m+\mu\1_\mathrm{m\text{ is used}},
\qquad 0<\mu<1.
\end{equation}
Consequently, persistent non-use decreases $\hat p_m$ and hence $U(m)$. Such items may be demoted, upon which their source experiences regain retrieval eligibility. UGC thus enables bidirectional cold--hot evolution under a fixed token budget rather than monotonic context growth.

\section{Experiments}
\subsection{Experimental Setup}
\label{sec:exp_setup}

\noindent\textbf{Benchmark and Execution.}
We evaluate SAGE on NPUKernelBench \citep{cao2026ascendkernelgen}, which
covers NPU kernel synthesis across multiple operator categories and
complexity levels. We extend the benchmark with iterative agent evaluation, per-round execution traces, and support for custom operators beyond the built-in tasks. All kernels are compiled and evaluated on Ascend 910C NPUs under a fixed CANN software stack. 

\noindent\textbf{Models and Budget.}
We use OpenCode \citep{opencode2026} as the agent harness and evaluate three source models spanning different scales: Qwen3-Coder-Next \citep{cao2026qwen3}, DeepSeek-V4-Flash \citep{xu2026deepseek} and GLM-5.2 \citep{zai2026glm52}. All methods share the same prompts, cold-start knowledge, CANN diagnostic tools, retrieval pipeline, and hardware evaluator.
We allow at most $T=30$ refinement rounds per operator, shared across its
correctness and optimization episodes.

\noindent\textbf{Metrics.}
Following NPUKernelBench, we report \textbf{Compilation Rate (CR)} and
\textbf{Execution Rate (ER)}, the fractions of operators for which a compilable
and functionally correct kernel is found within the interaction budget,
respectively. Let $\mathcal{O}$ denote the solved evaluation operators and
$T_{\mathrm{torch}}(o)$ the latency of the corresponding \texttt{torch\_npu}
implementation; $T_{\mathrm{first}}(o)$ and $T_{\mathrm{best}}(o)$ follow the
episode-level definitions in Sec.~\ref{sec:atu}. We report
\begin{equation}
\mathrm{Fast}_{1.0}
=\frac{1}{|\mathcal{O}|}\sum_{o\in\mathcal{O}}
\1_\mathrm{T_{\mathrm{torch}}(o)>T_{\mathrm{best}}(o)},
\qquad
S_{\mathrm{self}}
=\operatorname{median}_{o\in\mathcal{O}}
\frac{T_{\mathrm{first}}(o)}{T_{\mathrm{best}}(o)}.
\end{equation}
$\mathrm{Fast}_{1.0}$ measures the fraction of solved operators outperforming
their \texttt{torch\_npu} counterparts, while $S_{\mathrm{self}}$ measures performance gain achieved through iterative refinement.

\noindent\textbf{Baselines.}
Benchmark differences and limited access to complete implementations and knowledge bases hinder direct system-level comparisons with existing state-of-the-art methods. We therefore compare SAGE with three controlled baselines under the same OpenCode agent harness:
\begin{itemize}
    \item \textbf{Refinement} performs iterative repair and optimization within each operator but retains no cross-task experience;
    \item \textbf{Static RAG} uses the same initial cold-memory repository and retrieval pipeline as SAGE but disables subsequent memory writes and utility updates;
    \item \textbf{Value Memory} adapts EvoKernel's value-driven cross-task memory \citep{zheng2026towards} from its OpenReview supplementary material, using uniform credit assignment to give co-retrieved experiences the same verifier reward target for subsequent retrieval.
\end{itemize}  

\subsection{Main Results}
\label{sec:main_results}

NPUKernelBench exhibits a highly imbalanced distribution across operator
categories. We therefore construct a coverage-oriented subset of 88 operators:
20 Level-1 and 50 Level-2 tasks covering all categories at their respective levels, together with all 18 Level-3 tasks. Evaluation follows a sequential learning stream, with each task's result recorded before any memory updates for subsequent tasks. The complete operator list is
provided in Appendix~\ref{app:npu_operators}. Table~\ref{tab:main_results}
summarizes the results across three source models and four memory settings.

\begin{table}[H]
\caption{Main results on the 88-operator NPUKernelBench subset. All rates are
reported as percentage. The best result for each model is marked in bold.}
\label{tab:main_results}
\begin{center}
\renewcommand{\arraystretch}{1.10}
\setlength{\tabcolsep}{4pt}
\begin{tabular}{@{}llccccccc@{}}
\toprule
\multirow{2}{*}{\textbf{Model}} &
\multirow{2}{*}{\textbf{Method}} &
\multicolumn{3}{c}{\textbf{Execution Rate}} &
\multicolumn{2}{c}{\textbf{Overall}} &
\multicolumn{2}{c}{\textbf{Performance}} \\
\cmidrule(lr){3-5}
\cmidrule(lr){6-7}
\cmidrule(lr){8-9}
&
&
\textbf{L1} &
\textbf{L2} &
\textbf{L3} &
\textbf{CR} &
\textbf{ER} &
$\mathbf{Fast}_{1.0}$ &
$\mathbf{S}_{\mathrm{self}}$ \\
\midrule

\multirow{4}{*}{\shortstack[l]{Qwen3-Coder-Next (80B)}}
 & Refinement
 & 5.0 & 0.0 & 0.0
 & 35.2 & 1.1 & 0.0 & 1.00$\times$ \\
 & Static RAG
 & 30.0 & 26.0 & 0.0
 & 51.1 & 21.6 & 21.1 & 1.10$\times$ \\
 & Value Memory
 & 50.0 & 24.0 & \textbf{5.6}
 & 65.9 & 26.1 & 17.4 & 1.16$\times$ \\
 & SAGE
 & \textbf{55.0} & \textbf{32.0} & \textbf{5.6}
 & \textbf{75.0} & \textbf{31.8}
 & \textbf{28.6} & \textbf{1.27$\times$} \\
\midrule

\multirow{4}{*}{\shortstack[l]{DeepSeek-V4-Flash (284B)}}
 & Refinement
 & 35.0 & 18.0 & 0.0
 & 55.7 & 18.2 & 31.3 & 1.20$\times$ \\
 & Static RAG
 & 85.0 & 30.0 & 5.6
 & 73.9 & 37.5 & 42.4 & 1.42$\times$ \\
 & Value Memory
 & 80.0 & 58.0 & 27.8
 & 89.8 & 56.8 & 50.0 & 1.67$\times$ \\
 & SAGE
 & \textbf{90.0} & \textbf{62.0} & \textbf{44.4}
 & \textbf{95.5} & \textbf{64.8}
 & \textbf{64.9} & \textbf{2.08$\times$} \\
\midrule

\multirow{4}{*}{\shortstack[l]{GLM-5.2 (744B)}}
 & Refinement
 & 65.0 & 46.0 & 5.6
 & 79.5 & 42.0 & 48.6 & 1.74$\times$ \\
 & Static RAG
 & 75.0 & 60.0 & 38.9
 & 88.6 & 59.1 & 59.6 & 2.12$\times$ \\
 & Value Memory
 & 95.0 & 88.0 & 61.1
 & 97.7 & 84.1 & 75.7 & 3.36$\times$ \\
 & SAGE
 & \textbf{100.0} & \textbf{96.0} & \textbf{88.9}
 & \textbf{98.9} & \textbf{95.5}
 & \textbf{86.9} & \textbf{4.47$\times$} \\
\bottomrule
\end{tabular}
\end{center}
\end{table}

\noindent\textbf{Overall performance.}
SAGE achieves the strongest results across all three backbones and evaluation
dimensions. With GLM-5.2, it solves 84 of 88 operators within the interaction
budget, while most solved kernels outperform their corresponding
\texttt{torch\_npu} implementations. Relative to Value Memory,
$\mathrm{Fast}_{1.0}$ increases from 75.7\% to 86.9\%, while median
$S_{\mathrm{self}}$ improves from 3.36$\times$ to 4.47$\times$.
The gains persist on DeepSeek-V4-Flash and Qwen3-Coder-Next, indicating that
SAGE is not tied to a particular backbone.

\noindent\textbf{Effect of memory evolution.}
The progression from Refinement to Static RAG, Value Memory, and SAGE shows
consistent overall gains from increasingly structured cross-task experience
reuse, despite local non-monotonicity. The advantage becomes more pronounced
on harder tasks: SAGE raises Level-3 ER over Value Memory from 61.1\% to
88.9\% with GLM-5.2 and from 27.8\% to 44.4\% with DeepSeek-V4-Flash.
This widening gap suggests that reliable experience evaluation and
consolidation become increasingly important as task complexity grows.

\noindent\textbf{Backbone and optimization behavior.}
Backbone capability strongly affects how effectively accumulated knowledge is
exploited. Under SAGE, overall ER decreases from 95.5\% with GLM-5.2 to
64.8\% with DeepSeek-V4-Flash and 31.8\% with Qwen3-Coder-Next, with an even
larger separation on Level-3 tasks. GLM-5.2 nevertheless exhibits a long-tailed optimization distribution, with speedups ranging from \textbf{0.38$\times$ to
71.41$\times$}, \textbf{motivating the use of $\mathbf{Fast}_{1.0}$ together
with median $S_{\mathbf{self}}$} rather than arithmetic-mean speedup. Per-operator speedups relative to \texttt{torch\_npu} for the
correctness-passing Level-3/GMM tasks with GLM-5.2 are reported in
Appendix~\ref{app:lv3_operator_speedups}.
\subsection{Transfer Beyond the Benchmark}
\label{sec:xccl_transfer}

To examine whether the evolved memory remains effective beyond
NPUKernelBench, we further evaluate GLM-5.2 on 16 substantially more complex xCCL operators involving fused attention, normalization, cache management, and GMM-based computation. The complete operator list is provided in Appendix~\ref{app:xccl_operators}. Memory-based methods start from their post-benchmark memory states while retaining their native memory-update behavior, and all methods process the xCCL operators in the same fixed order under the same interaction budget. Thus, memory is allowed to continue evolving as the system adapts to the new operator library.
This evaluates adaptation to a new operator library from evolved memory, combining source-memory reuse with continued learning on xCCL.
\begin{figure}[H]
    \centering
    \includegraphics[width=0.73\linewidth]{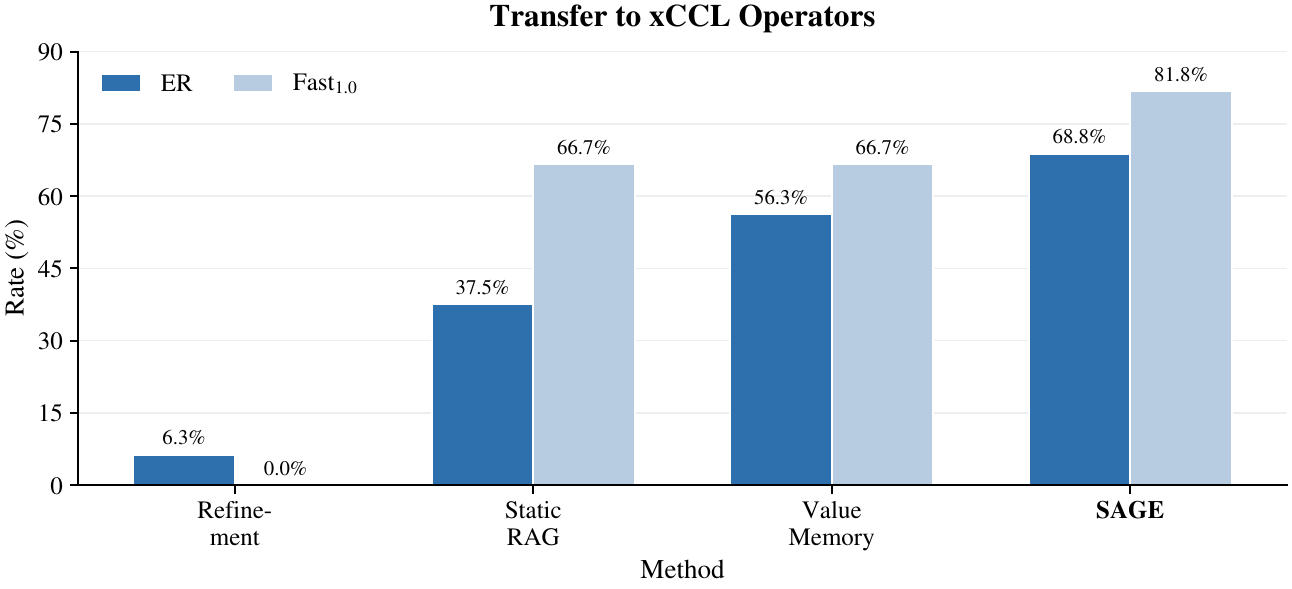}
    \caption{Transfer results on xCCL with GLM-5.2; memory updates remain enabled.}
    \label{fig:xccl_transfer}
\end{figure}
As shown in Figure~\ref{fig:xccl_transfer}, performance drops markedly for all methods relative to NPUKernelBench, reflecting the greater complexity and distribution shift of these operators. SAGE nevertheless solves 11 of 16 operators, achieving 68.8\% ER. Among these correct kernels, 81.8\% outperform their corresponding \texttt{torch\_npu} implementations. 

We further evaluate SAGE with GLM-5.3 on the same 16 xCCL operators, observing an ER increase from 68.8\% to 81.3\% and substantial performance gains across all operators solved by GLM-5.2. For sparse flash attention, a key kernel for efficient long-context LLM inference, speedup over \texttt{torch\_npu} increases from \textbf{23.35$\times$ to 43.99$\times$}. Its complete kernel code is provided as an example in Appendix~\ref{app:generated_kernel}. These results highlight SAGE's optimization potential on deployment-relevant kernels and provide further evidence that backbone capability substantially affects generated kernel performance, complementing the findings in Section~\ref{sec:main_results}. Per-operator results for both backbones are provided in Appendix~\ref{app:xccl_operator_speedups}.

\subsection{Effect of Task Order}
\label{app:task_order}

To examine how task order affects kernel synthesis, we use a fixed subset of 25 Level-2 operators as source tasks and all 18 Level-3 operators as targets. Then we compare three task schedules: 
\begin{itemize}
    \item \textbf{L3 Only}: SAGE processes only the Level-3 operator stream.
    \item \textbf{L2+L3 Mixed}: SAGE processes a fixed interleaving of the Level-2 and Level-3 operators.
    \item \textbf{L2$\rightarrow$L3}: SAGE first processes all Level-2 operators and then continues on the Level-3 stream using the resulting memory.
\end{itemize}

All schedules start from identical memory, use the same per-operator interaction budget, update memory after each task-solving episode, and never revisit operators. We report only Level-3 results, recorded before incorporating each operator's experience.

\begin{table}[H]
\centering
\caption{Cross-difficulty transfer to the 18 Level-3 operators. All rates are
percentages.}
\label{tab:cross_difficulty}
\renewcommand{\arraystretch}{1.08}
\setlength{\tabcolsep}{12pt}
\begin{tabular}{@{}lcc@{}}
\toprule
\textbf{Task schedule} & \textbf{L3 ER} & $\mathbf{Fast}_{1.0}$ \\
\midrule
L3 Only           & 38.9 & 42.9 \\
L2+L3 Mixed       & 61.1 & 63.6 \\
L2$\rightarrow$L3 & \textbf{77.8} & \textbf{71.4} \\
\bottomrule
\end{tabular}
\end{table}

The sequential \textit{L2$\rightarrow$L3} schedule achieves the strongest performance, improving L3 ER by 16.7 percentage points over the mixed stream and by 38.9 points over \textit{L3-only} adaptation. A consistent ordering is observed for $\mathrm{Fast}_{1.0}$. This progression indicates that consolidating lower-difficulty experience in advance provides a stronger basis for both functional synthesis and performance optimization on harder operators.

\subsection{Cross-Backbone Memory Transfer}
\label{sec:cross_backbone}

The backbone dependence observed in Section~\ref{sec:main_results} raises a
natural question: does SAGE accumulate model-specific experience, or can the
resulting knowledge be exploited by a different generator? We use GLM-5.2 as
the memory donor and evaluate DeepSeek-V4-Flash and Qwen3-Coder-Next under
three conditions: the shared initial memory before continual adaptation
(\textit{Initial}), memory evolved by the recipient itself (\textit{Native}),
and the complete SAGE memory evolved by GLM-5.2 on the same learning stream
(\textit{GLM Transfer}). Each condition independently reruns generation and hardware evaluation on the same 88 read-only operators with the same cold-start knowledge, diagnose tools and interaction budget; no memory updates are performed during
transfer evaluation. 
\begin{figure}[t]
    \centering
    \includegraphics[width=0.7\linewidth]{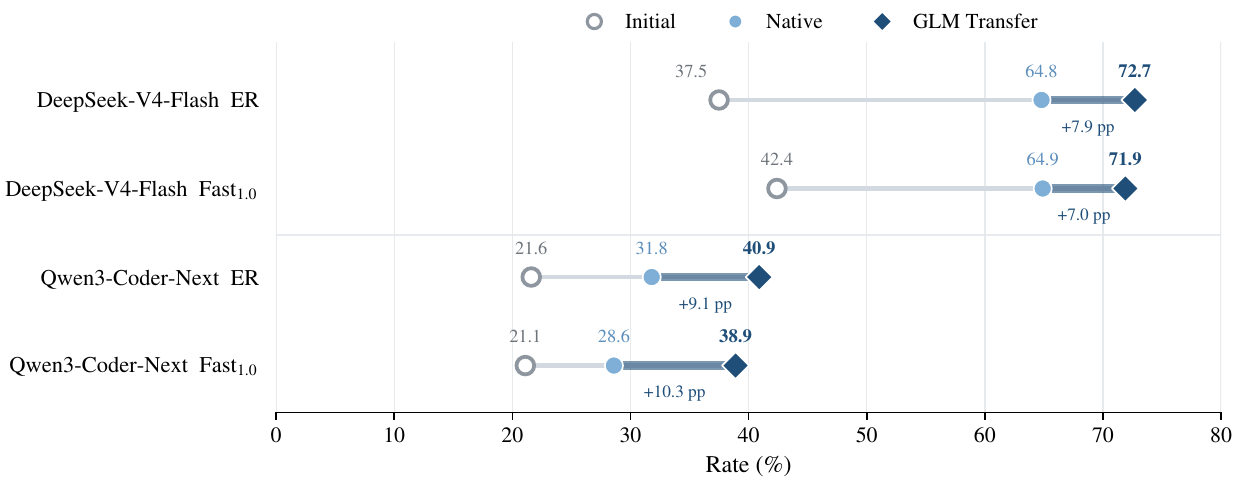}
    \caption{
    Cross-backbone transfer of initial, native-evolved, and GLM-5.2-evolved
    SAGE memory on the same read-only evaluation set.
    }
    \label{fig:cross_backbone}
\end{figure}

As shown in Figure~\ref{fig:cross_backbone}, both recipient backbones benefit
from memory evolved by GLM-5.2. For DeepSeek-V4-Flash, GLM Transfer raises ER
from 64.8\% with native memory to 72.7\%, while
$\mathrm{Fast}_{1.0}$ increases from 64.9\% to 71.9\%.
Qwen3-Coder-Next also improves, with ER increasing from 31.8\% to 40.9\%
and $\mathrm{Fast}_{1.0}$ from 28.6\% to 38.9\%.
These gains, obtained without parameter updates or additional memory
evolution, indicate that the knowledge accumulated by SAGE is partially
portable across backbone models. Nevertheless, the substantial performance
gap to GLM-5.2 remains, suggesting that transferred memory provides
complementary domain knowledge rather than compensating for differences in
backbone capability.

\subsection{Ablation Study}
\label{sec:ablation}

We ablate SAGE with GLM-5.2 on the same 88 operators. \textit{Uniform Credit} removes adoption tracing and assigns every retrieved experience the same terminal episode score, regardless of its adoption. \textit{w/o UGC} retains ATU and utility-aware retrieval but disables
cold--hot consolidation, while \textit{Random UGC} randomly selects eligible resident rules under the same token budget as SAGE.

\begin{figure}[H]
    \centering
    \includegraphics[width=0.7\linewidth]{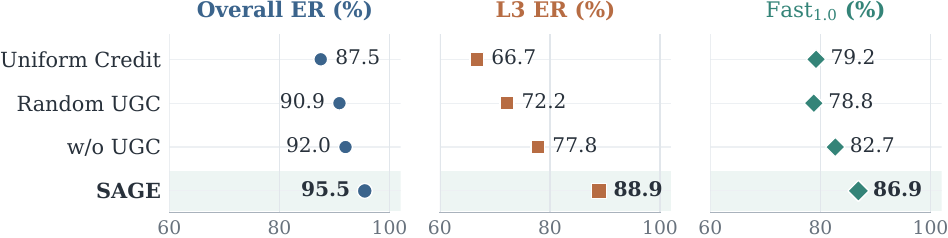}
    \caption{Ablation with GLM-5.2 on 88 operators. All metrics are percentages;
    $\mathrm{Fast}_{1.0}$ is conditional on correctness-passing operators.}
    \label{fig:ablation}
\end{figure}

Figure~\ref{fig:ablation} shows complementary benefits from adoption-aware credit assignment and utility-guided consolidation. Uniform Credit yields the largest L3 ER drop (88.9\% to 66.7\%). This suggests that coarse-grained attribution is especially
harmful on complex kernel synthesis trajectories. Removing UGC retains much of the benefit of ATU but remains below full SAGE, while Random UGC performs worse than utility-guided consolidation. Together, these results indicate that SAGE benefits not merely from accumulating more context, but from identifying useful experience and selectively converting it into persistent knowledge.

\section{Conclusion}
We presented SAGE, a persistent self-improving agent for NPU kernel synthesis that enables frozen LLMs to accumulate hardware-specific expertise through external memory evolution. Rather than treating retrieved experiences uniformly, SAGE traces experience adoption to estimate item-level utility and consolidates repeatedly validated knowledge under a bounded hot-memory budget. Across three backbones, SAGE improves both correctness and efficiency, with larger gains on
complex Level-3 operators; transfer results and mechanism ablations support the role of accurate attribution and selective consolidation. However, our 30-round study does not cover long-horizon synthesis of fused kernels and megakernels, which requires diversity-preserving search across coupled scheduling, memory-layout, and synchronization decisions. In practice, human kernel programmers may use isolated SAGE sandboxes or jointly update shared cold and hot memories, raising issues of specialization, provenance, transfer, and conflict resolution.
Future work will explore this principle across other emerging hardware
architectures and domain-specific programming languages.

\bibliography{references}
\bibliographystyle{unsrt}

\appendix

\section{Evaluation Operator Lists}
\label{app:eval_operators}
\subsection{NPUKernelBench Evaluation Subset}
\label{app:npu_operators}

Table~\ref{tab:eval_operators} lists the 88 NPUKernelBench operators used in our main evaluation, comprising all Level-3 tasks and coverage-oriented subsets of Levels 1 and 2 from the full benchmark.

\begin{table}[H]
\caption{Operators included in the 88-task NPUKernelBench evaluation subset.}
\label{tab:eval_operators}
\begin{center}
\renewcommand{\arraystretch}{1.08}
\begin{tabular}{p{0.08\textwidth} p{0.13\textwidth} p{0.70\textwidth}}
\toprule
\textbf{Level} & \textbf{Category} & \textbf{Operators} \\
\midrule

\multirow{5}{*}{Level 1}
& Comparison
& Equal, Less, LessEqual \\
& Condition
& IsFinite, IsInf, NonFiniteCheckOp \\
& Index
& GatherV3, ScatterList \\
& Math
& Sqrt, SinMath, CastMath, AbsMath, Addcdiv, ClipByValue, CosMath, Lerp,
StridesliceNegConcatV2 \\
& TensorCreation
& Arange, Eye, Fill \\

\midrule

\multirow{9}{*}{Level 2}
& Activation
& DequantSwigluQuant, FastGelu, Gelu, InplaceAttnSoftmax, SwiGlu, Swish,
GeGluV2, GeluGrad, MulSigmoid \\
& Foreach
& ForeachAbs, ForeachAddScalar, ForeachMulScalar, ForeachDivScalar,
ForeachExp, ForeachSqrt, ForeachNeg, ForeachZeroInplace, ForeachAddList,
ForeachMulList, ForeachSubScalar, ForeachCos, ForeachSigmoid \\
& Linalg
& Cross \\
& Loss
& CrossEntropyLoss, MseLoss, CrossEntropyLossGrad, MseLossGradV2 \\
& Mask
& MaskedSelectV3, Tril, Triu \\
& Norm
& LayerNormV4, RmsNorm, DeepNorm, BatchNormV3, GroupNormSwish,
AddLayerNorm, PreLayerNorm, AddRmsNorm, InplaceAddRmsNorm,
GroupNormSwishGrad \\
& Optim
& ApplyAdamWV2, ApplyFusedEmaAdam \\
& Reduce
& AddSigmoidMulReduceSumD, SelectReduceMaxDSubExpReduceSumDRealDiv,
MulMulReduceMeanDTwice, MulSigmoidMulAddCustom \\
& TensorMove
& CoalesceSparse, StridedSliceAssignV2, ExpandV2, ReverseSequence \\

\midrule

\multirow{2}{*}{Level 3}
& GMM
& BasicMatmul, BatchedMatmul, FlashAttentionScoreWithLargeHeadDim, Gemm,
Gemv, GroupGemm, GroupedMatmulSliceK,
GroupedMatmulSliceKPerTokenDequant, GroupedMatmulSliceM,
GroupedMatmulSliceMPerTokenDequant, MatmulAdd, MatmulBias, Mla,
OptimizedMatmul, PaddingMatmul, QuantMatmul, SplitkMatmul \\
& Sort
& TopKV3 \\

\bottomrule
\end{tabular}
\end{center}
\end{table}
\subsection{xCCL Transfer Evaluation Set}
\label{app:xccl_operators}

Table~\ref{tab:xccl_operators} lists the 16 xCCL operators used for the
cross-library transfer evaluation in Section~\ref{sec:xccl_transfer}. 
% These operators cover substantially more complex fused computation patterns,
% including attention, normalization, cache management, quantization, and
% GMM-based computation.

\begin{table}[H]
\caption{The 16 xCCL operators used in the transfer evaluation.}
\label{tab:xccl_operators}
\centering
\renewcommand{\arraystretch}{1.05}
\setlength{\tabcolsep}{6pt}
\begin{tabular}{@{}ll@{}}
\toprule
\textbf{Operator} & \textbf{Operator} \\
\midrule
lightning\_indexer
&
sparse\_flash\_attention\\
mla\_prolog\_v3\_x
&
compute\_and\_cache\_digest\\
apply\_rotary\_pos\_emb\_partial
&
gmm\_deq\_swiglu\_quant\_gmm\_deq\\
hc\_post\_x

&
hc\_post\\
fused\_gelu\_sigmoid
&
fused\_scatter\_update\_and\_compute\_digest\\
rms\_norm\_rope
&
grouped\_matmul\_swiglu\_quant\_x
\\
moe\_dist\_dispatch\_mul\_exp
&
moe\_gating\_top\_k\_remap\\
rotary\_pos\_emb\_partial
&
dequant\_swiglu\_clamp\_quant\\
\bottomrule
\end{tabular}
\end{table}

\subsection{Level-3/GMM Per-Operator Speedups}
\label{app:lv3_operator_speedups}

\begin{figure}[H]
  \centering
  \includegraphics[width=0.75\linewidth]{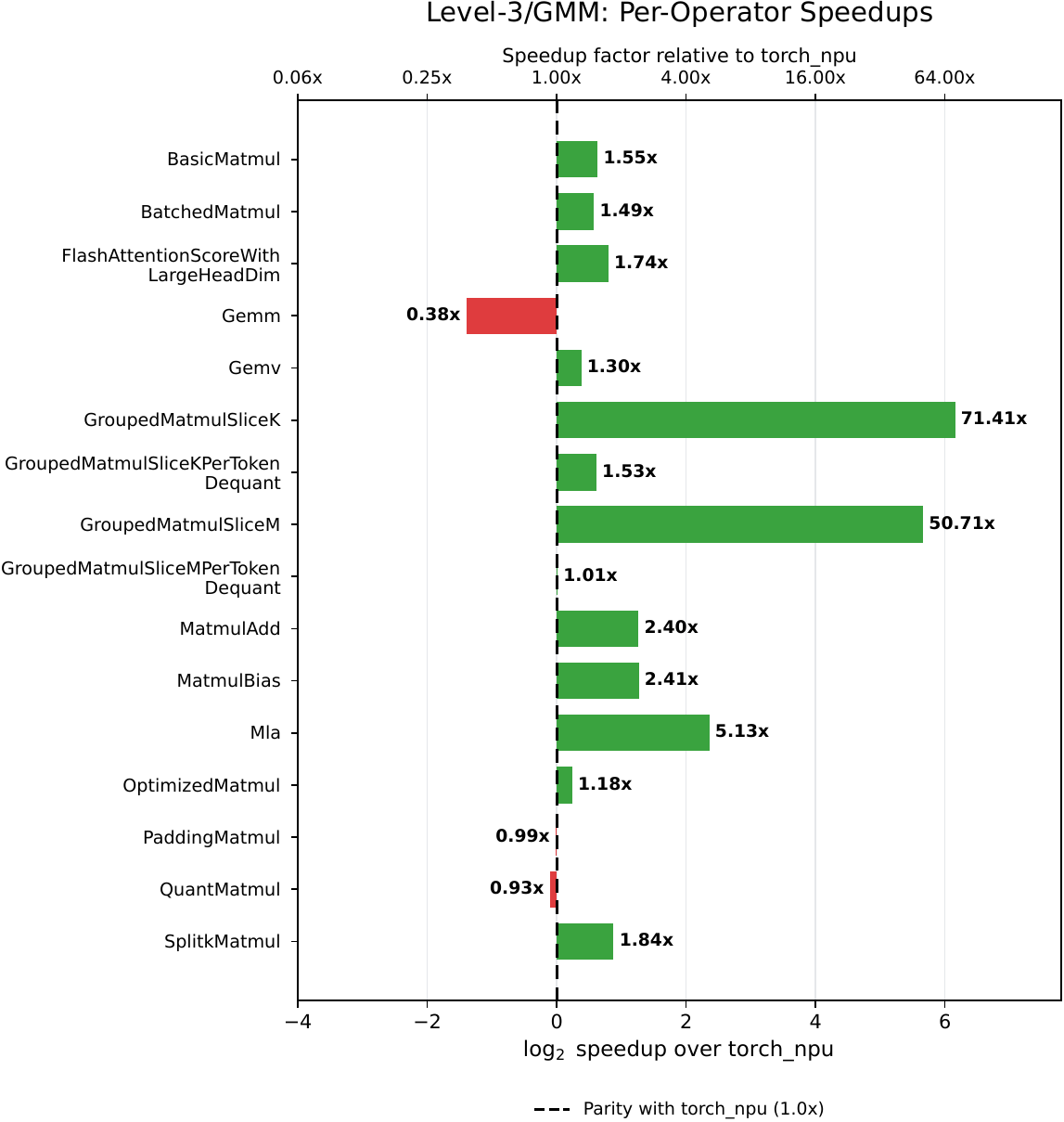}
  \caption{Per-operator speedups relative to \texttt{torch\_npu} for
  correctness-passing Level-3/GMM tasks with GLM-5.2. Green and red indicate
  speedups above and below $1\times$, respectively; the dashed line marks
  parity. Horizontal distances are log$_2$-transformed.}
  \label{fig:lv3_operator_speedups}
\end{figure}

Figure~\ref{fig:lv3_operator_speedups} reports the best correct SAGE kernel's
speedup over \texttt{torch\_npu} for each of the 16 correctness-passing
Level-3/GMM operators. The median speedup is $1.54\times$; individual speedups span $0.38\times$ to $71.41\times$, illustrating why a single aggregate cannot
describe the optimization distribution. Bars show log$_2$ speedup, while their
annotations retain the original multiplicative values.

\subsection{xCCL Per-Operator Speedups}
\label{app:xccl_operator_speedups}
\begin{figure}[H]
  \centering
  \includegraphics[width=0.8\linewidth]
  {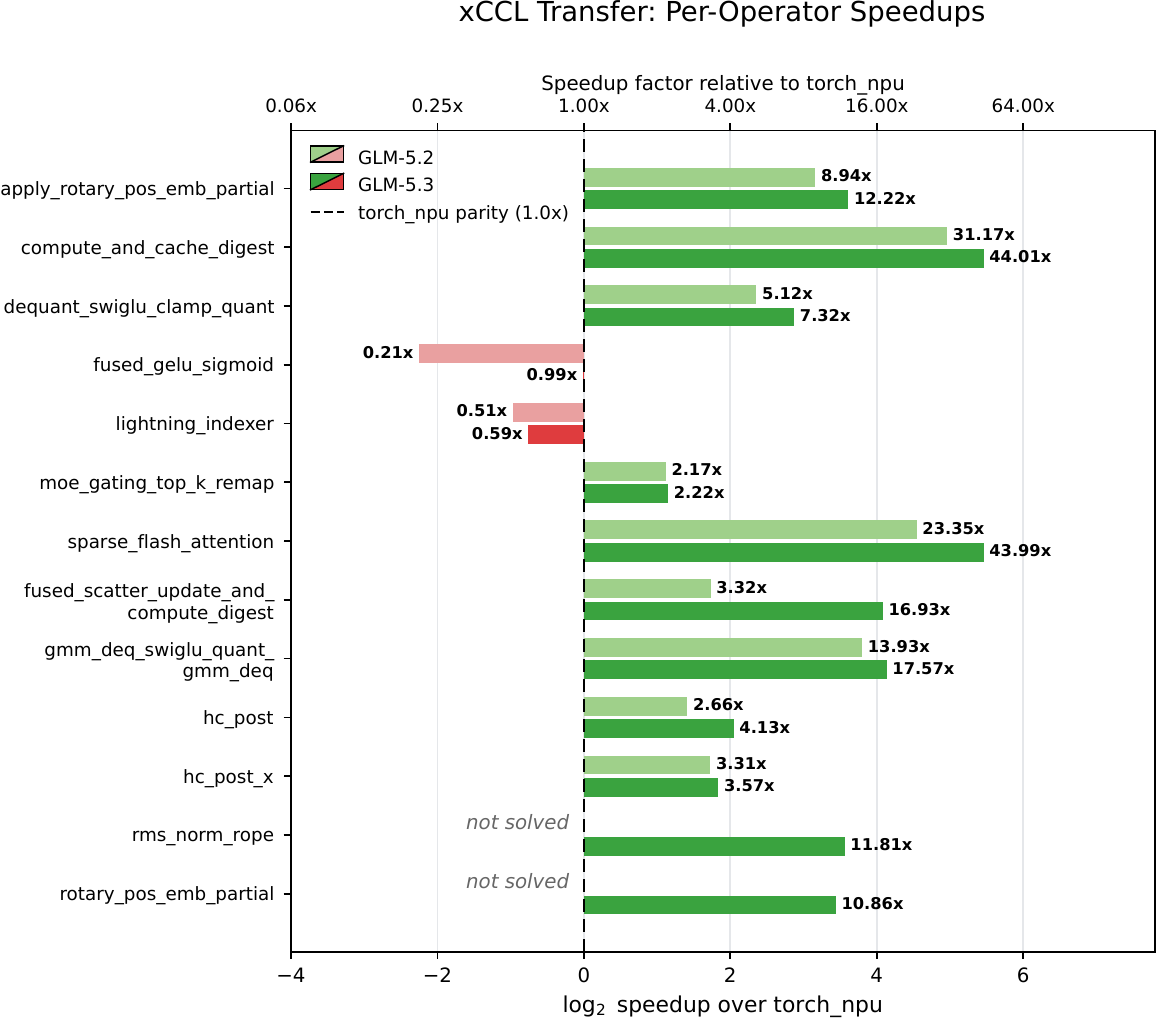}
  \caption{Per-operator speedups relative to \texttt{torch\_npu} for xCCL
  tasks solved by GLM-5.2 or GLM-5.3. Paired bars compare the two backbones
  on the 11 commonly solved operators. The horizontal scale and color convention match Figure~\ref{fig:lv3_operator_speedups}.}
  \label{fig:xccl_operator_speedups}
\end{figure}

Figure~\ref{fig:xccl_operator_speedups} reports per-operator speedups for
the xCCL tasks solved by either GLM-5.2 or GLM-5.3. On the 11 operators
solved by both backbones, the median speedup over \texttt{torch\_npu}
increases from 3.32$\times$ with GLM-5.2 to 7.32$\times$ with GLM-5.3.
GLM-5.3 additionally solves \texttt{rms\_norm\_rope} and
\texttt{rotary\_pos\_emb\_partial}, achieving 11.81$\times$ and
10.86$\times$ speedups, respectively; its median over all 13
correctness-passing xCCL operators is 10.86$\times$. Both evaluations
inherit memory evolved over all 88 NPUKernelBench tasks, providing a
mature starting point for cross-library adaptation.

\section{Hyperparameter Configuration}
\label{app:hyperparameters}
Table~\ref{tab:hyperparameters} summarizes the main hyperparameters used by
SAGE. Unless otherwise stated, the same configuration is used across all
backbones and evaluation sets.
\begin{table}[t]
\centering
\caption{Main hyperparameters used throughout SAGE.}
\label{tab:hyperparameters}
\renewcommand{\arraystretch}{1.08}
\setlength{\tabcolsep}{6pt}
\begin{tabular}{@{}llll@{}}
\toprule
\textbf{Symbol} & \textbf{Value} & \textbf{Component} & \textbf{Description} \\
\midrule
$T$                  & 30         & Agent loop & Maximum interaction rounds \\
$\delta$             & 0.2        & ATU        & Unsuccessful-episode penalty \\
$\beta$              & 0.2        & ATU        & Unadopted-retrieval penalty \\
$\eta_{\min}$        & 0.05       & ATU        & Minimum utility update rate \\
$\gamma$             & 0.3        & Retrieval  & Utility reranking strength \\
$K_{\mathrm{ado}}$   & 3          & UGC        & Minimum adopted trajectories \\
$K_{\mathrm{op}}$    & 3          & UGC        & Minimum distinct operators \\
$B$                  & 10K tokens & UGC        & Resident hot-memory budget \\
$\mu$                & 0.1        & UGC        & Hot-memory usage EMA coefficient \\
\bottomrule
\end{tabular}
\end{table}

We use moderate negative credit, setting $\delta=\beta=0.2$, so that a
single failed episode or retrieved-but-unadopted item does not dominate
accumulated positive evidence. Utility-aware reranking uses $\gamma=0.3$,
keeping semantic relevance as the primary retrieval signal while allowing
historical utility to influence ranking. For consolidation, we set
$K_{\mathrm{ado}}=K_{\mathrm{op}}=3$, requiring repeated adoption across
multiple operators before an experience becomes eligible for promotion.
The resident hot memory is bounded by $B=10$K tokens.

Hyperparameters are selected once based on their intended numerical roles
and preliminary system-level stability. The same configuration is used for
Qwen3-Coder-Next, DeepSeek-V4-Flash, and GLM-5.2; no per-backbone or
per-operator tuning is performed.

\section{System Implementation and Evaluation Protocol}
\label{app:implementation}

\subsection{Utility-Gated Consolidation Procedure}
\label{app:ugc_procedure}

Algorithm~\ref{alg:ugc} details the budgeted replacement step of UGC, which is
executed after each task-solving episode. UGC pools the current hot-memory
items $\mathcal{H}$ with validated candidates $\mathcal{V}$ that satisfy the
admission gates, ranks them by expected utility per resident token $U(m)$, and
reconstructs the hot region under budget $B$. An item that does not fit is
skipped while the scan continues, allowing shorter lower-ranked items to be
admitted. Newly admitted items become consolidated, whereas displaced
residents revert to validated status and regain retrieval eligibility.

\begin{algorithm}[H]
\caption{Budgeted Hot-Memory Replacement}
\label{alg:ugc}
\begin{algorithmic}[1]
\Require Current hot memory $\mathcal{H}$, validated candidates
$\mathcal{V}$, utility density $U(\cdot)$, token budget $B$
\Ensure Updated hot memory $\mathcal{H}'$

\State $\mathcal{P} \gets \mathcal{H} \cup \mathcal{V}$
\State $\mathcal{H}' \gets \varnothing$
\ForAll{$m \in \mathcal{P}$ sorted by decreasing $U(m)$}
\If{$\operatorname{Tok}(\mathcal{H}') + L_m \le B$}
\State $\mathcal{H}' \gets \mathcal{H}' \cup {m}$
\EndIf
\EndFor

\State $\sigma_m \gets \mathrm{consolidated},
\ \forall m\in\mathcal{H}'\setminus\mathcal{H}$
\State $\sigma_m \gets \mathrm{validated},
\ \forall m\in\mathcal{H}\setminus\mathcal{H}'$

\State \Return $\mathcal{H}'$
\end{algorithmic}
\end{algorithm}

\subsection{Infrastructure and Agent Runtime}
\label{app:system_runtime}
\paragraph{Execution environment.}
Experiments are executed on Ascend 910C infrastructure with
8 accelerator cards (16 accelerator dies in total) per server. Multiple
kernel-generation workers run concurrently, with each agent worker assigned
one accelerator card. The host environment is Huawei Cloud EulerOS 2.0 on
ARM64. The core software stack is:

\begin{itemize}
    \item \textbf{NPU runtime:} CANN 8.5.1 with driver 25.5.1;
    \item \textbf{Python stack:} Python 3.11.14,
    \texttt{torch\_npu} 2.9.0, and \texttt{vllm-ascend} 0.18.0rc1;
    \item \textbf{Model serving:} all evaluated backbones are deployed locally
    through \texttt{vllm-ascend}. DeepSeek-V4-Flash uses
    DP$=4$/TP$=4$, while GLM-5.2 and GLM-5.3 use TP$=16$; both enable expert parallelism
    and Ascend-native quantization. Qwen3-Coder-Next follows the official
    \texttt{vllm-ascend} deployment recipe.
\end{itemize}

Unspecified serving parameters follow the corresponding
\texttt{vllm-ascend} defaults. The agent runtime is implemented through
OpenCode skills, which orchestrate the iterative generation, compilation,
execution, diagnosis, and memory-update loop; the evaluated local backbone
provides the kernel-generation and reasoning capability.

\paragraph{Device-level diagnostics.}
The agent can autonomously invoke both lightweight AscendC probes and
CANN-native debugging/profiling utilities:

\begin{itemize}
    \item \textbf{In-kernel probes:}
    \texttt{printf}, \texttt{DumpTensor}, and \texttt{DumpAccChkPoint};
    \item \textbf{Runtime debugging:}
    \texttt{msDebug} and \texttt{msSanitizer} for memory violations,
    race conditions, synchronization failures, and device-side faults;
    \item \textbf{Performance diagnosis:}
    \texttt{msProf} and \texttt{torch\_npu.profiler} for kernel-level
    execution and bottleneck analysis.
\end{itemize}

These tools are exposed to the agent as callable diagnostics rather than
manually injected feedback, enabling tool-augmented localization of runtime
and performance failures during refinement.

\subsection{Initialization and Experimental Integrity}
\label{app:experimental_integrity}

\paragraph{Cold-start knowledge construction.}
SAGE initializes architecture-specific knowledge through document distillation and differential mining. We distill and manually consolidate official AscendC API documentation, operator-development tool user guides, and AscendC best practices into three lightweight references: (i) a compact AscendC quick reference, (ii) a debugging guide, and (iii) an \texttt{add\_custom} implementation example. Conditioned on these references, the agent generates kernels for simple Level-1 and Level-2 NPUKernelBench operators. Once compilation succeeds, we compare the generated kernel with an expert solution to extract reusable implementation patterns for cold retrieval memory. The operators used for differential mining are strictly disjoint from the 88 evaluation operators.

\paragraph{Controlled comparison and evaluation protocol.}
All memory-based methods start from an identical initial memory state. Refinement, Static RAG, Value Memory, and SAGE share the same backbone, agent harness, prompt structure, diagnostic tool access, interaction budget ($T=30$), and evaluation harness; only the memory mechanism differs. The 88 evaluation operators form the learning stream, with each result recorded before any corresponding memory updates for subsequent operators. Cross-backbone transfer evaluates the same operators using fixed, previously evolved memories, without further memory updates.

\paragraph{Anti-hacking safeguards.}
During task solving on evaluation operators, an autonomous coding agent may
exploit implementation shortcuts that inflate apparent correctness or speedup
without reflecting genuine kernel-synthesis capability. We therefore enforce
the following evaluation-integrity constraints for all methods:

\begin{itemize}
    \item \textbf{No equivalent high-level substitution.}
    The agent may not implement the target computation by directly invoking a
    higher-level AscendC primitive that subsumes its core functionality (e.g., using \texttt{AscendC::Sigmoid} as the core
    implementation of Swish).
    \item \textbf{No expert-implementation leakage.}
    The agent may not inspect or search expert or structurally similar operator
    implementations outside the designated example directory.
    \item \textbf{No semantic shortcuts.}
    Optimizations may not remove required functionality, omit boundary
    handling, or hard-code tiling parameters for specific evaluation inputs.
    \item \textbf{No invalid precision trade-offs.}
    Reduced-precision transformations are permitted only when the resulting
    kernel satisfies the same correctness criteria used for evaluation.
\end{itemize}

\subsection{Correctness and Performance Measurement}
\label{app:evaluation_protocol}

\paragraph{Correctness validation.}
For NPUKernelBench operators, generated kernels are validated against the
corresponding \texttt{torch\_npu} implementation using the benchmark error
metrics, including average/maximum absolute error, average/maximum relative
error, and average accuracy. A kernel is considered correct when
\begin{equation}
\mathrm{max\_abs\_error}\leq10^{-3},
\qquad
\mathrm{max\_rel\_error}\leq10^{-3}.
\end{equation}

\paragraph{Latency measurement.}
Latency is measured using device-side Ascend NPU events to avoid host-side
scheduling and synchronization overhead in wall-clock timing. We instantiate
the event timer with \texttt{torch.npu.Event(enable\_timing=True)} and apply
the same measurement protocol to every generated kernel and its
\texttt{torch\_npu} reference:

\begin{itemize}
    \item execute 5 warm-up iterations;
    \item synchronize the NPU and record the start event;
    \item execute 100 timed iterations;
    \item record the end event, synchronize, and compute elapsed event time.
\end{itemize}

We report the mean per-execution latency over the 100 timed iterations.
Generated kernels and their \texttt{torch\_npu} references use identical
inputs, warm-up counts, repetition counts, and synchronization boundaries.
For each operator,
\[
T_{\mathrm{best}}
=
\min_{k\in\mathcal K_{\mathrm{correct}}} T(k),
\]
where $\mathcal K_{\mathrm{correct}}$ denotes all correctness-passing
candidates produced within the interaction budget.

\section{Additional Analyses and Controlled Experiments}
\label{app:controlled_experiments}
\subsection{Expert Audit of Adoption Records}
\label{app:adoption_audit}

\noindent\textbf{Audit scope.}
We examine whether recorded adoption corresponds to observable code revisions. We randomly sample 50 records from all records labeled adopted in the saved generation traces. The sampling unit is an adoption record.

\noindent\textbf{Verification.}
For each sampled record, an Ascend C expert among the authors checks the associated experience and adoption rationale against the corresponding code revision, assigning one of three correspondence ratings: \emph{full}, \emph{partial}, or \emph{unrelated}. The saved trajectory provides the code evidence for this assessment.

\noindent\textbf{Result and interpretation.}
The expert rated 44 records (88\%) as full, six (12\%) as partial, and none as unrelated (Table~\ref{tab:adoption_audit}). Every inspected record thus showed at least partial correspondence with code revisions. These findings support recorded adoption as a trace-grounded proxy for actual memory use, providing an empirical basis for adoption-aware credit assignment in ATU.

\begin{table}[H]
\centering
\caption{Expert-rated correspondence between adoption and code revisions for 50 records randomly sampled from all trace records labeled adopted.}
\label{tab:adoption_audit}
\begin{tabular}{@{}lcc@{}}
\toprule
\textbf{Correspondence} & \textbf{Count} & \textbf{Percentage} \\
\midrule
Full      & 44 & 88\% \\
Partial   & 6  & 12\% \\
Unrelated & 0  & 0\% \\
\bottomrule
\end{tabular}
\end{table}
\subsection{Fixed-Bank Credit Assignment}
\label{app:fixed_bank}

The end-to-end ablation in Section~\ref{sec:ablation} changes the memory
accumulated over time together with its utility estimates. To isolate credit
quality itself, we freeze an identical final SAGE memory bank, including its
cold and hot contents, and replay the same historical adoption traces to
construct three cold-memory ranking variants: relevance-only retrieval
($\gamma=0$), uniform credit that assigns the terminal score to every retrieved
item, and ATU. Hot memory remains identical across all variants, so
only cold-memory utility reranking differs.

\begin{table}[H]
\centering
\caption{Credit-assignment control under a shared frozen memory bank.
All rates are percentages.}
\label{tab:fixed_bank}
\renewcommand{\arraystretch}{1.08}
\setlength{\tabcolsep}{8pt}
\begin{tabular}{@{}lccc@{}}
\toprule
\textbf{Ranking signal} & \textbf{ER} & \textbf{L3 ER} &
$\mathbf{Fast}_{1.0}$ \\
\midrule
Relevance Only & 84.1 & 61.1 & 70.3 \\
Uniform Credit   & 89.8 & 72.2 & 79.7 \\
ATU    & \textbf{95.5} & \textbf{88.9} & \textbf{86.9} \\
\bottomrule
\end{tabular}
\end{table}

Under the same memory contents, ATU consistently improves both correctness
and optimization over uniform credit and relevance-only ranking. ATU consistently outperforms both controls across all reported metrics, supporting the intended role of adoption tracing: verifier
outcomes provide aggregate supervision over generated kernels, whereas ATU
more precisely attributes this evidence to the retrieved experiences actually
used during synthesis.

\section{Formal Properties of ATU and UGC}
\label{app:formal_properties}

We establish several properties of ATU and UGC that motivate their design
choices in persistent kernel synthesis. The analysis focuses on bounded credit,
adaptive utility estimation, controlled utility-aware retrieval, and
usage-aware consolidation.

\subsection{Property 1: Bounded and normalized adoption credit.}
\label{app:property_1}
Recall that the terminal episode score satisfies $z\in[-\delta,1]$, and ATU
assigns
\begin{equation}
\xi_m =
\begin{cases}
z/|\mathcal A|, & m\in\mathcal A,\\
-\beta, & m\in\mathcal R\setminus\mathcal A,
\end{cases}
\label{eq:app_credit}
\end{equation}
where $\mathcal R$ and $\mathcal A$ are the episode-deduplicated retrieved and
adopted sets, respectively. Let $q=\max\{\delta,\beta\}$ and assume
$0\leq\delta,\beta\leq1$. Then
\begin{equation}
\xi_m\in[-q,1],
\qquad
\sum_{m\in\mathcal A}\xi_m=z
\quad
(\mathcal A\neq\varnothing).
\label{eq:app_credit_properties}
\end{equation}

\emph{Proof.}
For $m\in\mathcal A$, $|\mathcal A|\geq1$ and
$z/|\mathcal A|\in[-\delta,1]$; otherwise $\xi_m=-\beta$, proving the first
claim. For the second,
$\sum_{m\in\mathcal A}z/|\mathcal A|=z$.
\hfill$\square$

Thus, the total credit assigned to adopted experiences equals the terminal
episode score regardless of how many experiences are adopted. This normalization prevents larger adoption sets from receiving more total
credit. Moreover, moderate $\delta$ and $\beta$ are
appropriate for kernel synthesis: an unsuccessful candidate does not imply
that every adopted hardware insight was harmful. With our default
$\delta=\beta=0.2$, negative item-level credit is lower bounded by $-0.2$.

\subsection{Property 2: Stable but continually adaptive utility.}
\label{app:property_2}
ATU updates each item according to
\begin{equation}
u_m \leftarrow
(1-\eta_m)u_m+\eta_m\xi_{m},
\qquad
\eta_m=
\max\left\{
\frac{1}{1+n_m^{\mathrm{ret}}},
\eta_{\min}
\right\},
\label{eq:app_utility}
\end{equation}
with $u_m^{(0)}=0$ and $0<\eta_{\min}\leq1$.
If $\xi_m\in[-q,1]$, then
\begin{equation}
u_m^{(t)}\in[-q,1]
\qquad \forall t.
\label{eq:app_utility_bound}
\end{equation}

\emph{Proof.}
Since $0<\eta_m\leq1$, Eq.~\eqref{eq:app_utility} is a convex combination of
$u_m$ and $\xi_m$. The interval $[-q,1]$ is convex and contains the
initial value, so the result follows by induction.
\hfill$\square$

The reciprocal term progressively stabilizes an item as retrieval evidence
accumulates, while the nonzero floor prevents its utility from becoming
permanently frozen. Starting from the utility estimate at update $t$, after $h$ subsequent updates, the weight retained from $u_m^{(t)}$ is at most
\begin{equation}
\prod_{j=0}^{h-1}(1-\eta_{m,t+j})
\leq
(1-\eta_{\min})^h.
\label{eq:app_forgetting}
\end{equation}
Hence old evidence is forgotten geometrically. For
$\eta_{\min}=0.05$, this bound corresponds to a half-life of approximately
$13.5$ updates.
This behavior is intentional: unlike stationary bandit estimation, the utility
of kernel experience can change as the task stream, available memory, and
generator context evolve. ATU therefore tracks retrieval-conditioned utility
rather than requiring convergence to a fixed stationary value.

\subsection{Property 3: Controlled utility-aware reranking.}
\label{app:property_3}
Historical utility should help distinguish among semantically plausible
experiences, but should not make a previously useful yet task-irrelevant item
dominate retrieval. SAGE therefore uses utility only as a multiplicative
modulation of the relevance score:
\begin{equation}
s(m)=s_{\mathrm{rel}}(m)(1+\gamma u_m),
\qquad
s_{\mathrm{rel}}(m)\geq0,\quad 0\leq\gamma<1.
\label{eq:app_reranking}
\end{equation}
From Property~\ref{app:property_2}, $u_m\in[-q,1]$, where
$q=\max\{\delta,\beta\}\leq1$. It follows directly that
\begin{equation}
1-\gamma q
\leq
1+\gamma u_m
\leq
1+\gamma,
\label{eq:app_modulation}
\end{equation}
so the modulation factor remains strictly positive. 

We next characterize when utility can change the ordering between two
candidates. Consider $m_1$ and $m_2$ with positive relevance scores, and
suppose $m_1$ is semantically more relevant,
$s_{\mathrm{rel}}(m_1)>s_{\mathrm{rel}}(m_2)$. The most adversarial utility
configuration for preserving this ordering assigns the minimum possible
utility to $m_1$ and the maximum possible utility to $m_2$, i.e.,
$u_{m_1}=-q$ and $u_{m_2}=1$. Therefore,
\begin{equation}
s(m_1)\geq
s_{\mathrm{rel}}(m_1)(1-\gamma q),
\qquad
s(m_2)\leq
s_{\mathrm{rel}}(m_2)(1+\gamma).
\end{equation}
Hence a sufficient condition for the relevance ordering to remain unchanged
under \emph{any} admissible utility values is
\begin{equation}
\frac{s_{\mathrm{rel}}(m_1)}
{s_{\mathrm{rel}}(m_2)}
>
\frac{1+\gamma}{1-\gamma q}
\quad\Longrightarrow\quad
s(m_1)>s(m_2).
\label{eq:relevance_dominance}
\end{equation}

With our defaults $q=0.2$ and $\gamma=0.3$,
\begin{equation}
1+\gamma u_m\in[0.94,1.30],
\qquad
\frac{1+\gamma}{1-\gamma q}\approx1.38.
\end{equation}
Thus utility can reorder candidates with comparable semantic relevance, while
sufficiently large relevance gaps remain preserved. The asymmetry is intentional: positive utility reflects successful adoption-traced evidence, whereas negative evidence is treated conservatively because non-adoption or episode failure does not necessarily imply harmful knowledge. This is particularly appropriate for NPU kernel synthesis, where a historically useful experience may still be inapplicable to a structurally different operator.

\subsection{Property 4: Resident priority decays under persistent non-use.}
\label{app:property_4}
UGC updates the usage estimate of each resident item as
\begin{equation}
\hat p_m
\leftarrow
(1-\mu)\hat p_m
+ \mu\1_\mathrm{m\text{ is used}},
\qquad 0<\mu<1,
\label{eq:app_usage_ema}
\end{equation}
and ranks resident knowledge by
\begin{equation}
U(m)=\frac{u_m\hat p_m}{L_m}.
\label{eq:app_resident_density}
\end{equation}
For analysis, let $\hat p_m^{(t)}$ denote the usage estimate after the
$t$-th memory update. If item $m$ is not used during the next $h$ updates,
Eq.~\eqref{eq:app_usage_ema} recursively gives
\begin{equation}
\hat p_m^{(t+h)}
=
(1-\mu)^h\hat p_m^{(t)}.
\label{eq:app_usage_decay}
\end{equation}
If $u_m$ and $L_m$ remain unchanged over this interval, its resident priority
decays at the same rate:
\begin{equation}
U_{t+h}(m)
=
(1-\mu)^h U_t(m).
\label{eq:app_density_decay}
\end{equation}

The decay rate has a direct temporal interpretation. The number of consecutive
unused updates required to halve the usage contribution is
\begin{equation}
h_{1/2}
=
\frac{\log(1/2)}{\log(1-\mu)}.
\label{eq:app_usage_halflife}
\end{equation}
With our default $\mu=0.1$, $h_{1/2}\approx6.6$ memory updates. Thus recently
useful knowledge is retained smoothly, while persistently unused knowledge
loses resident priority rather than remaining permanently consolidated.

More importantly, this decay guarantees finite-time loss of priority relative
to any persistently useful alternative. Consider a stale resident item $m_a$
with $U_t(m_a)>0$, and suppose there exists an alternative candidate $m_b$
whose priority satisfies
\[
U_{t+j}(m_b)\geq c>0
\qquad \text{for all } j\geq0,
\]
where $c<U_t(m_a)$. If $m_a$ remains unused and its utility and token cost
remain unchanged, then
\begin{equation}
U_{t+h}(m_a) = (1-\mu)^h U_t(m_a) < c
\end{equation}
for any integer $h$ satisfying
\begin{equation}
h > \frac{\log\!\left(U_t(m_a)/c\right)} {-\log(1-\mu)}.
\label{eq:app_finite_displacement}
\end{equation}
Since $0<1-\mu<1$, the right-hand side is finite. Hence a persistently useful candidate eventually outranks any resident item that remains unused. 

This property is central to UGC: consolidation is not irreversible promotion.
In persistent NPU kernel synthesis, knowledge useful for earlier operators may become less relevant as computation patterns, tensor layouts, or optimization regimes change. Usage-aware decay retains recurrently useful knowledge while allowing stale expertise to lose resident priority.

\section{Qualitative Examples of Memory Evolution and Kernel Synthesis}
\label{app:qualitative_examples}

This appendix complements the aggregate evaluation with concrete examples of how SAGE selects retrieved experience, assigns post-episode credit, represents evolving memory, and produces complete Ascend~C kernels. We first examine two task-solving trajectories that expose the distinction between retrieval and actual adoption. We then document the serialized memory schemas and provide a representative fused kernel generated by SAGE, together with the task-level instruction used to invoke the skill-based agent.

\subsection{Adoption-Aware Memory Use}
\label{app:qualitative_adoption}

\noindent\textbf{Diagnosing a host--device boundary error in GatherV3.}
At iteration 9, \texttt{GatherV3} failed during correctness validation with a
segmentation fault. The preceding revision had removed \texttt{ValueDepend}
from the scalar \texttt{axis} input, but the host-side tiling function still
dereferenced the data pointer returned by \texttt{GetData}. Without
\texttt{ValueDepend}, this call exposes an NPU device address that cannot be
dereferenced by the host CPU. Table~\ref{tab:gatherv3_adoption} summarizes the
five retrieved experiences and the agent's adoption decisions.

\begin{table}[H]
\centering
\caption{Retrieved experiences for the GatherV3 correctness-stage runtime
failure. Relevance is the retriever score; adoption is determined from the
subsequent diagnosis and code revision.}
\label{tab:gatherv3_adoption}
\setlength{\tabcolsep}{5pt}
\renewcommand{\arraystretch}{1.08}
\begin{tabularx}{\textwidth}{@{}cYcc@{}}
\toprule
\textbf{ID} & \textbf{Retrieved experience} & \textbf{Rel.} & \textbf{Adopted} \\
\midrule
181 & Host-side tiling cannot dereference device-resident tensor data returned
by \texttt{GetData}. & 69.17 & Yes \\
193 & A kernel entry can read a scalar directly from GM through a
\texttt{\_\_gm\_\_} pointer cast. & 67.42 & Yes \\
61 & Scalar inputs can be exposed to tiling through
\texttt{ValueDepend(REQUIRED)} and passed in \texttt{TilingData}. & 57.43 & No \\
139 & Nested tiling structures require
\texttt{GET\_TILING\_DATA\_WITH\_STRUCT}. & 49.22 & No \\
159 & \texttt{TilingData} field types must match their kernel-side consumers. &
32.47 & No \\
\bottomrule
\end{tabularx}
\end{table}

The agent adopted items 181 and 193: the former identified the failure mechanism,
whereas the latter supplied the applicable repair. Guided by the adopted pair, the agent removed all axis-dependent computation from the host tiling function, retained only input dimensions in \texttt{TilingData}, and
moved the scalar read and dependent work partitioning into the device side:

\begin{lstlisting}[style=sagecpp,backgroundcolor=\color{sagekernelbg},numbers=none]
int64_t axisVal = *reinterpret_cast<__gm__ int64_t*>(axis);
if (axisVal < 0) 
    axisVal += static_cast<int64_t>(dimNum);
uint32_t axis = static_cast<uint32_t>(axisVal);
this->srcDimSize = xDims[axis];

uint32_t outerSize = 1;
for (uint32_t d = 0; d < axis; ++d)
    outerSize *= xDims[d];
uint32_t innerSize = 1;
for (uint32_t d = axis + 1; d < dimNum; ++d)
    innerSize *= xDims[d];
\end{lstlisting}

This example illustrates why relevance alone is insufficient for memory credit:
all five items were retrieved for the same failure, yet only two were causally
reflected in the repair.

\medskip
\noindent\textbf{Kernel synthesis trajectories of Mla.}
The Level-3 \texttt{Mla} (Multi-head Latent Attention) operator of NPUKernelBench fuses two QK paths, softmax, RoPE, and PV matrix multiplication. SAGE required 23 generation attempts to obtain the first correct kernel and then used the remaining budget for optimization.
Table~\ref{tab:mla_trace} condenses the principal state transitions.

\begin{table}[t]
\centering
\caption{Key transitions in the Mla task-solving trajectory on
\texttt{ascend910\_93}.}
\label{tab:mla_trace}
\setlength{\tabcolsep}{2.2pt}
\renewcommand{\arraystretch}{1.08}
\begin{tabularx}{\textwidth}{@{}p{0.1\textwidth}YYp{0.18\textwidth}@{}}
\toprule
\textbf{Iteration} & \textbf{Diagnosis} & \textbf{Revision} & \textbf{Outcome} \\
\midrule
0--2 & A shared Matmul object changed its $K$ dimension despite fixed tiling;
the default task type also incorrectly assigned Cube work on AIV core. & Separate the QK
and PV configurations and declare
\texttt{KERNEL\_TYPE\_MIX\_AIC\_1\_0}. & Compilation succeeds; Matmul no
longer hangs. \\
10--13 & Multi-token execution exposed unmatched AIC--AIV synchronization. &
Isolate a single-token path, then pair cross-core events within the token loop. &
The hang becomes a diagnosable numerical mismatch. \\
18--23 & \texttt{DumpTensor} showed zero softmax inputs; four host inputs were
non-contiguous, and the QK path omitted the RoPE dimensions. & Synchronize the
asynchronous Matmul write, apply \texttt{AutoContiguous()}, and materialize the
full 576-dimensional Q/K representations. & First correct kernel that provides  optimization baseline: $0.420$ ms ($1.91\times$). \\
24--26 & Double buffering violated the MTE3 handoff; a global barrier was also
overly conservative. & Revert the unsafe buffer change and replace
\texttt{PIPE\_ALL} with an MTE3--MTE2 event. & $1.95\times$. \\
27 & Work remained single-core. & Partition query tokens across cores. &
$3.71\times$. \\
29 & A relay copy materialized contiguous QK data before PV. & Split PV into
nope and RoPE Matmuls that read the original GM tensors directly. & $0.155$ ms;
$5.13\times$. \\
\bottomrule
\end{tabularx}
\end{table}

% At the initial retrieval, the agent adopted three items addressing tiling-header API access and Matmul shape management. It rejected the high-scoring task-type suggestions because they did not explain the failure at that point; a task-type correction became necessary only in a later iteration. The shape-consistency constraint in item 166 remained reflected in the first correct Mla kernel. By the end of the trajectories, the item had been recorded as adopted for BasicMatmul, GroupGemm, and Mla. This cross-operator evidence allows its lifecycle state to advance to \texttt{validated}, while subsequent consolidation remains subject to UGC's utility gate.

\subsection{Memory Record Schemas}
\label{app:memory_schemas}

Tables~\ref{tab:item_record_schema} and~\ref{tab:hot_snapshot_schema} document SAGE's item records and hot-memory snapshots. Each item record stores a retrieval-facing rule, diagnostic evidence, ATU utility, retrieval and adoption counts, cross-operator adoption history, and lifecycle state. These fields remain in the cold store after promotion, preserving the item's history. The Example values are synthetic and illustrate the schemas only.

A hot-memory snapshot identifies the items selected under a resident token budget. Before promotion, access frequency is estimated from retrievals over eligible episodes. Resident items then bypass retrieval, and their access estimates are updated from observed context usage. Their utility per token, \(U(m)=u_m\hat p_m/L_m\), combines estimated benefit, access frequency, and context cost.

\begin{table}[H]
\centering
\caption{Schema of a memory-item record.}
\label{tab:item_record_schema}
\setlength{\tabcolsep}{4pt}
\small
\renewcommand{\arraystretch}{1.07}
\begin{tabularx}{\textwidth}{@{}p{0.26\textwidth}p{0.11\textwidth}Yp{0.26\textwidth}@{}}
\toprule
\textbf{Field} & \textbf{Type} & \textbf{Description} & \textbf{Example} \\
\midrule
\texttt{id} & integer & Stable memory-item identifier. & \texttt{10035} \\
\texttt{title} & string & Compact retrieval-facing rule. & Reduce algorithmic complexity and branches \\
\texttt{type} & enum & Experience type used for filtering. & \texttt{algorithmic\_optimization} \\
\texttt{category} & string & Operator or failure category. & \texttt{Scan/Reduction} \\
\texttt{error\_message} & string & Failure or inefficiency that motivated the item. & Quadratic polynomial evaluation with redundant writes \\
\texttt{code\_diff} & object & Minimal incorrect/correct code pair. & \texttt{\{wrong\_code, correct\_code\}} \\
\texttt{summary} & string & Compressed explanation stored for retrieval. & Replace repeated powers with Horner evaluation \\
\texttt{u\_m} & float & ATU utility estimate. & \texttt{0.42} \\
\texttt{n\_ret} & integer & Number of retrievals. & \texttt{6} \\
\texttt{n\_ado} & integer & Number of observed adoptions. & \texttt{4} \\
\texttt{adopted\_operators} & list[string] & Distinct operators providing adoption evidence. & \texttt{[GroupGemm, ...]} \\
\texttt{sigma} & enum & Lifecycle state. & \texttt{consolidated} \\
\texttt{L\_m} & integer & Resident token cost. & \texttt{96} \\
\texttt{hot\_region} & boolean & Whether the item currently resides in hot memory. & \texttt{true} \\
\texttt{hot\_last\_used\_episode} & integer/null & Most recent resident-context use. & \texttt{18} \\
\texttt{p\_hat} & float & Estimated access frequency; updated by usage EMA after promotion. & \texttt{0.35} \\
\texttt{n\_elig} & integer & Episodes for which the cold item was retrieval-eligible. & \texttt{18} \\
\texttt{density} & float & Expected utility per resident token, $U(m)$. & \texttt{0.00153} \\
\bottomrule
\end{tabularx}
\end{table}

\begin{table}[t]
\centering
\caption{Schema of a hot-memory snapshot and its nested entry representation.}
\label{tab:hot_snapshot_schema}
\setlength{\tabcolsep}{4pt}
\small
\renewcommand{\arraystretch}{1.07}
\begin{tabularx}{\textwidth}
{@{}p{0.20\textwidth}p{0.14\textwidth}Yp{0.22\textwidth}@{}}
\toprule
\textbf{Field} & \textbf{Type} & \textbf{Description} & \textbf{Example} \\
\midrule
\texttt{session\_id} & integer & Memory-evolution session identifier. & \texttt{1} \\
\texttt{created\_at} & timestamp & Snapshot creation time. & \texttt{20260811\_111049} \\
\texttt{budget} & integer & Maximum resident-context tokens. & \texttt{10000} \\
\texttt{n\_ops} & integer & Completed operator trajectories at snapshot time. & \texttt{18} \\
\texttt{hot\_entries} & list[object] & Items selected under the token budget. & \texttt{[\{id: 238, ...\}]} \\
\midrule
\multicolumn{4}{@{}l}{\emph{Nested hot-entry fields}} \\
\texttt{id} & integer & Identifier of the resident item. & \texttt{238} \\
\texttt{title} & string & Consolidated rule inserted into persistent context. & Use int32 generation plus an int64 cast for ArithProgression \\
\texttt{c\_tokens} & integer & Tokens consumed by the consolidated representation. & \texttt{61} \\
\texttt{u} & float & Current ATU utility estimate. & \texttt{0.62} \\
\texttt{p\_hat} & float & Resident-context usage estimate. & \texttt{0.44} \\
\texttt{density} & float & Ranking score $U(m)=u_m\hat p_m/L_m$. & \texttt{0.00447} \\
\bottomrule
\end{tabularx}
\end{table}

\subsection{Generated Kernel and Skill-Based Task Instruction}
\label{app:generated_kernel}

SAGE is implemented through executable skills rather than a monolithic
handcrafted prompt. The generator skill constructs and validates a complete
Ascend~C operator project; after correctness is reached, the optimizer skill
continues from the best feasible kernel using on-device performance feedback.
The task-level instruction used across experiments follows the template below.

\begin{lstlisting}[style=sageinstruction,linewidth=\linewidth]
Use the ascend-kernel-generator skill to generate the {operator_category} operator {operator_name} until it passes compilation and correctness tests, and then use the ascend-kernel-optimizer skill for continual optimization. You may execute at most 30 iterations in total; once this budget is reached, stop immediately regardless of progress. Follow the complete skill workflows and do not use any hacking shortcuts. Code references are restricted to the skill's own examples directory, and examples for other operators must not be inspected. When an error occurs, proactively apply the debugging techniques provided in the skill references to localize the faulty code. The local SocVersion is ascend910_93.
\end{lstlisting}

We present the sparse flash attention kernel generated by SAGE using GLM-5.3 as its backbone during the xCCL evaluation. It passed correctness validation and achieved a
$43.99\times$ speedup over the reference \texttt{torch\_npu} implementation.
The file-specific background colors distinguish host registration and tiling
(blue), the serialized tiling definition (yellow), and device-kernel code
(green).

\noindent\textbf{\texttt{op\_host/sparse\_flash\_attention.cpp}}
\begin{lstlisting}[style=sagecpp,backgroundcolor=\color{sagehostbg}]
#include "sparse_flash_attention_tiling.h"
#include "register/op_def_registry.h"
#include "tiling/platform/platform_ascendc.h"
#include "tiling/tiling_api.h"

using namespace matmul_tiling;

namespace optiling {
// kernel-side task granularity: gSize q-heads joined (each q-head -> 1 row batch
// of gSize contiguous q rows sharing the same sparse selection).
static constexpr int32_t S1_TILE = 1;
static constexpr int32_t BASE_M = 16;
static constexpr int32_t BASE_N = 16;
static constexpr int32_t BASE_K = 64;
// kernel-side select-loop granularity (queries per select loop, <= S1_TILE rows
// are re-selected for each new query row; kept a host constant so kernel and
// tiling agree).
static constexpr int32_t SEL_ROWS = 1;
static constexpr int32_t N_BATCH = 32;
// MLA: 512 nope dims + 64 rope dims; output carries nope cols only.
static constexpr int32_t NOPE_DIM = 512;

static ge::graphStatus TilingFunc(gert::TilingContext *context)
{
    auto ascendcPlatform = platform_ascendc::PlatformAscendC(context->GetPlatformInfo());
    auto qShape = context->GetInputShape(0)->GetStorageShape();
    uint32_t B = static_cast<uint32_t>(qShape.GetDim(0));
    uint32_t S1 = static_cast<uint32_t>(qShape.GetDim(1));
    uint32_t N1 = static_cast<uint32_t>(qShape.GetDim(2));

    auto kShape = context->GetInputShape(1)->GetStorageShape();
    // key layout: PA_BSND [blockNum, blockSize, N2, D]
    uint32_t P = static_cast<uint32_t>(kShape.GetDim(1));
    uint32_t N2 = static_cast<uint32_t>(kShape.GetDim(2));

    auto idxShape = context->GetInputShape(3)->GetStorageShape();
    // sparse_indices: [B, N2, S1, K]
    uint32_t K = static_cast<uint32_t>(idxShape.GetDim(3));

    uint32_t gSize = (N2 == 0) ? N1 : N1 / N2;

    // head-group split: gSize % HG == 0 -> tasks = B*S1*(gSize/HG); each task
    // runs the full gather+bmm1+softmax+bmm2 chain for HG joined rows of one
    // (b, s1), enabling multi-core across the head dimension.
    uint32_t HG = 16;  // 8 and 16 both pass with relays fixed (same 0.106ms);
                       // keep 16: fewer tasks, larger bmm M per task.
    while (HG > 1 && (gSize % HG != 0)) {
        HG /= 2;
    }
    uint32_t nGroups = gSize / HG;

    uint32_t aicNum = ascendcPlatform.GetCoreNumAic();
    // one task = one (b, s1, headGroup): total joined-row batches to schedule
    uint32_t totalTasks = B * S1 * nGroups;
    uint32_t usedCoreNum = (totalTasks < aicNum) ? totalTasks : aicNum;
    if (usedCoreNum == 0) {
        usedCoreNum = 1;
    }

    uint64_t ubSizePlat = 0;
    ascendcPlatform.GetCoreMemSize(platform_ascendc::CoreMemType::UB, ubSizePlat);
    uint64_t l1SizePlat = 0;
    ascendcPlatform.GetCoreMemSize(platform_ascendc::CoreMemType::L1, l1SizePlat);
    int32_t l1Size = static_cast<int32_t>(l1SizePlat);

    SparseFlashAttentionTilingData tiling;
    tiling.set_batch(B);
    tiling.set_seqQ(S1);
    tiling.set_numHeadsQ(N1);
    tiling.set_blockSize(P);
    tiling.set_sparseBlkSize(P);   // sparse block granularity == page granularity
    tiling.set_sparseBlkCount(K);
    tiling.set_gSize(gSize);
    tiling.set_hgSplit(HG);
    tiling.set_nIdxSize((K + N_BATCH - 1) / N_BATCH);

    auto attrs = context->GetAttrs();
    float scaleValue = 1.0f;
    if (attrs != nullptr) {
        const float *scalePtr = attrs->GetFloat(0);
        if (scalePtr != nullptr) {
            scaleValue = *scalePtr;
        }
    }
    tiling.set_scaleValue(scaleValue);

    uint32_t S2s = K * P;  // selected KV span (columns of score matrix)

    // bmm1: [HG, S2s, 576]  A=Qjoined(BSND row-major, non-trans)
    //        B=KVgathered^T (gathered KV cache laid out [S2s, 576] rows, transposed)
    MultiCoreMatmulTiling cubeTiling1(ascendcPlatform);
    cubeTiling1.SetDim(1);
    cubeTiling1.SetAType(TPosition::GM, CubeFormat::ND, matmul_tiling::DataType::DT_BF16, false);
    cubeTiling1.SetBType(TPosition::GM, CubeFormat::ND, matmul_tiling::DataType::DT_BF16, true);
    cubeTiling1.SetCType(TPosition::GM, CubeFormat::ND, matmul_tiling::DataType::DT_FLOAT);
    cubeTiling1.SetShape(HG, S2s, 576);
    cubeTiling1.SetOrgShape(gSize, S2s, 576);
    cubeTiling1.SetFixSplit(BASE_M, BASE_N, BASE_K);
    cubeTiling1.SetBias(false);
    cubeTiling1.SetBufferSpace(l1Size, -1, -1);
    if (cubeTiling1.GetTiling(tiling.cubeTiling1) == -1) {
        return ge::GRAPH_FAILED;
    }

    // bmm2: [HG, 512, S2s]  A=P(joined row-major), B=Vgathered nope cols
    // (non-trans); C width 512 matches the [B,S1,N1,512] output row stride.
    MultiCoreMatmulTiling cubeTiling2(ascendcPlatform);
    cubeTiling2.SetDim(1);
    cubeTiling2.SetAType(TPosition::GM, CubeFormat::ND, matmul_tiling::DataType::DT_BF16, false);
    cubeTiling2.SetBType(TPosition::GM, CubeFormat::ND, matmul_tiling::DataType::DT_BF16, false);
    cubeTiling2.SetCType(TPosition::GM, CubeFormat::ND, matmul_tiling::DataType::DT_BF16);
    cubeTiling2.SetShape(HG, NOPE_DIM, S2s);
    cubeTiling2.SetOrgShape(gSize, NOPE_DIM, S2s);
    cubeTiling2.SetFixSplit(BASE_M, BASE_N, BASE_K);
    cubeTiling2.SetBias(false);
    cubeTiling2.SetBufferSpace(l1Size, -1, -1);
    if (cubeTiling2.GetTiling(tiling.cubeTiling2) == -1) {
        return ge::GRAPH_FAILED;
    }

    // softmax over S2s: src [SOFTMAX_M, S2s] half
    constexpr uint32_t SOFTMAX_M = 4;
    std::vector<int64_t> shapeVec = {SOFTMAX_M, static_cast<int64_t>(S2s)};
    ge::Shape softmaxSrcShape(shapeVec);
    uint32_t softmaxMinTmp = AscendC::GetSoftMaxMinTmpSize(softmaxSrcShape, sizeof(float), false);
    uint32_t softmaxMaxTmp = AscendC::GetSoftMaxMaxTmpSize(softmaxSrcShape, sizeof(float), false);
    uint32_t localWorkSpaceSize = softmaxMaxTmp;
    AscendC::SoftMaxTilingFunc(softmaxSrcShape, sizeof(float), localWorkSpaceSize, tiling.softmaxTiling);
    (void)softmaxMinTmp;

    tiling.set_usedCoreNum(usedCoreNum);
    tiling.set_blockDim(usedCoreNum);
    tiling.SaveToBuffer(context->GetRawTilingData()->GetData(), context->GetRawTilingData()->GetCapacity());
    context->GetRawTilingData()->SetDataSize(tiling.GetDataSize());

    context->SetBlockDim(usedCoreNum);

    // workspace: sys part + 16MB reserved gap + user part
    // user per block: score [g, S2s] float | P [g, S2s] half | KVg [S2s,576] bf16
    //                  | Vg [S2s,512] bf16 | Qj [g,576] bf16
    constexpr uint64_t RESERVED_WS = 16ULL * 1024 * 1024;
    uint64_t scoreSlot = static_cast<uint64_t>(HG) * S2s * sizeof(float);
    uint64_t pSlot = static_cast<uint64_t>(HG) * S2s * sizeof(half);
    uint64_t kvSlot = static_cast<uint64_t>(S2s) * 576 * 2;
    uint64_t vgSlot = static_cast<uint64_t>(S2s) * NOPE_DIM * 2;
    uint64_t qjSlot = static_cast<uint64_t>(HG) * 576 * 2;
    uint64_t userWorkspaceSize = (scoreSlot + pSlot + kvSlot + vgSlot + qjSlot) * usedCoreNum;
    size_t systemWorkspaceSize = static_cast<size_t>(ascendcPlatform.GetLibApiWorkSpaceSize());
    size_t *currentWorkspace = context->GetWorkspaceSizes(1);
    currentWorkspace[0] = userWorkspaceSize + systemWorkspaceSize + RESERVED_WS;

    return ge::GRAPH_SUCCESS;
}
} // namespace optiling

namespace ge {
static ge::graphStatus InferShape(gert::InferShapeContext *context)
{
    const gert::Shape *queryShape = context->GetInputShape(0);
    gert::Shape *attentionOutShape = context->GetOutputShape(0);
    *attentionOutShape = *queryShape;
    return ge::GRAPH_SUCCESS;
}

static ge::graphStatus InferDataType(gert::InferDataTypeContext *context)
{
    context->SetOutputDataType(0, context->GetInputDataType(0));
    return ge::GRAPH_SUCCESS;
}
} // namespace ge

namespace ops {
class SparseFlashAttention : public OpDef {
public:
    explicit SparseFlashAttention(const char* name) : OpDef(name)
    {
        this->Input("query")
            .ParamType(REQUIRED)
            .DataType({ge::DT_BF16})
            .Format({ge::FORMAT_ND})
            .UnknownShapeFormat({ge::FORMAT_ND});
        this->Input("key")
            .ParamType(REQUIRED)
            .DataType({ge::DT_BF16})
            .Format({ge::FORMAT_ND})
            .UnknownShapeFormat({ge::FORMAT_ND});
        this->Input("value")
            .ParamType(REQUIRED)
            .DataType({ge::DT_BF16})
            .Format({ge::FORMAT_ND})
            .UnknownShapeFormat({ge::FORMAT_ND});
        this->Input("sparse_indices")
            .ParamType(REQUIRED)
            .DataType({ge::DT_INT32})
            .Format({ge::FORMAT_ND})
            .UnknownShapeFormat({ge::FORMAT_ND});
        this->Input("block_table").ParamType(OPTIONAL).DataType({ge::DT_INT32})
            .Format({ge::FORMAT_ND}).UnknownShapeFormat({ge::FORMAT_ND});
        this->Input("actual_seq_lengths_query").ParamType(OPTIONAL).DataType({ge::DT_INT32})
            .Format({ge::FORMAT_ND}).UnknownShapeFormat({ge::FORMAT_ND});
        this->Input("actual_seq_lengths_kv").ParamType(OPTIONAL).DataType({ge::DT_INT32})
            .Format({ge::FORMAT_ND}).UnknownShapeFormat({ge::FORMAT_ND});
        this->Input("query_rope").ParamType(OPTIONAL).DataType({ge::DT_BF16})
            .Format({ge::FORMAT_ND}).UnknownShapeFormat({ge::FORMAT_ND});
        this->Input("key_rope").ParamType(OPTIONAL).DataType({ge::DT_BF16})
            .Format({ge::FORMAT_ND}).UnknownShapeFormat({ge::FORMAT_ND});
        this->Output("attention_out")
            .ParamType(REQUIRED)
            .DataType({ge::DT_BF16})
            .Format({ge::FORMAT_ND})
            .UnknownShapeFormat({ge::FORMAT_ND});
        this->Attr("scale_value").AttrType(OPTIONAL).Float(1.0);
        this->Attr("sparse_block_size").Int();
        this->Attr("layout_query").AttrType(OPTIONAL).String("BSND");
        this->Attr("layout_kv").AttrType(OPTIONAL).String("PA_BSND");
        this->Attr("sparse_mode").AttrType(OPTIONAL).Int(3);

        this->SetInferShape(ge::InferShape).SetInferDataType(ge::InferDataType);
        this->AICore()
                .SetTiling(optiling::TilingFunc);
        this->AICore().AddConfig("ascend910_93")
                      .AddConfig("ascend910b");
    }
};
OP_ADD(SparseFlashAttention);
} // namespace ops
\end{lstlisting}

\noindent\textbf{\texttt{op\_host/sparse\_flash\_attention\_tiling.h}}
\begin{lstlisting}[style=sagecpp,backgroundcolor=\color{sagetilingbg}]
#ifndef SPARSE_FLASH_ATTENTION_TILING_H
#define SPARSE_FLASH_ATTENTION_TILING_H

#include "register/tilingdata_base.h"
#include "tiling/tiling_api.h"

namespace optiling {
BEGIN_TILING_DATA_DEF(SparseFlashAttentionTilingData)
TILING_DATA_FIELD_DEF(uint32_t, batch);            // B
TILING_DATA_FIELD_DEF(uint32_t, seqQ);             // S1 rows of query per batch
TILING_DATA_FIELD_DEF(uint32_t, numHeadsQ);        // N1
TILING_DATA_FIELD_DEF(uint32_t, blockSize);        // KV page size P
TILING_DATA_FIELD_DEF(uint32_t, sparseBlkSize);    // sparse block size (== blockSize in our tests)
TILING_DATA_FIELD_DEF(uint32_t, sparseBlkCount);   // K = selected blocks per (b,hq,s)
TILING_DATA_FIELD_DEF(uint32_t, gSize);            // N1 / N2 (N2 == 1)
TILING_DATA_FIELD_DEF(uint32_t, hgSplit);          // q-heads per task group (gSize % hgSplit == 0)
TILING_DATA_FIELD_DEF(uint32_t, nIdxSize);         // ceil(K / N_BATCH): per-select-loop UB rows
TILING_DATA_FIELD_DEF(uint32_t, usedCoreNum);      // number of AIC:AIV blocks launched
TILING_DATA_FIELD_DEF(uint32_t, blockDim);
TILING_DATA_FIELD_DEF(float, scaleValue);          // softmax scale
TILING_DATA_FIELD_DEF_STRUCT(TCubeTiling, cubeTiling1); // bmm1: [S1_TILE, S2s, 576]
TILING_DATA_FIELD_DEF_STRUCT(TCubeTiling, cubeTiling2); // bmm2: [S1_TILE, 576, S2s]
TILING_DATA_FIELD_DEF_STRUCT(SoftMaxTiling, softmaxTiling); // softmax over S2s, src half
END_TILING_DATA_DEF;

REGISTER_TILING_DATA_CLASS(SparseFlashAttention, SparseFlashAttentionTilingData)
} // namespace optiling

#endif
\end{lstlisting}

\noindent\textbf{\texttt{op\_kernel/sparse\_flash\_attention.cpp}}
\begin{lstlisting}[style=sagecpp,backgroundcolor=\color{sagekernelbg}]
#include "kernel_operator.h"
#include "lib/matmul_intf.h"

using namespace AscendC;
using namespace matmul;

// MLA sparse attention decode: per query row, gather sparse-selected KV pages
// (nope 512 + rope 64 = 576 cols), bmm1(Qjoined x KVg^T) -> score float,
// softmax -> P half, bmm2(P x KVg) -> O bf16. KFC MIX AIC 1:2, AIV drives cube.
constexpr uint32_t SOFTMAX_ROWS = 4;
constexpr uint32_t FLOAT_NUM_PER_BLOCK = 8;
constexpr uint32_t NOPE_DIM = 512;
constexpr uint32_t ROPE_DIM = 64;
constexpr uint32_t DQK = NOPE_DIM + ROPE_DIM;  // 576
constexpr uint32_t N_BATCH = 32;               // sparse indices resolved per loop
constexpr uint32_t P_MAX = 16;                 // max KV page rows staged per relay

using AT1 = MatmulType<TPosition::GM, CubeFormat::ND, bfloat16_t, false>;  // Qj [g, 576]
using BT1 = MatmulType<TPosition::GM, CubeFormat::ND, bfloat16_t, true>;   // KVg [S2s, 576] trans
using CT1 = MatmulType<TPosition::GM, CubeFormat::ND, float>;              // score float
using AT2 = MatmulType<TPosition::GM, CubeFormat::ND, bfloat16_t, false>;  // P  [g, S2s] bf16
using BT2 = MatmulType<TPosition::GM, CubeFormat::ND, bfloat16_t, false>;  // Vg [S2s, 576]
using CT2 = MatmulType<TPosition::GM, CubeFormat::ND, bfloat16_t>;         // out bf16

// REGIST_MATMUL_OBJ must stay in the extern-C entry scope (KfcServer/KfcCommClient
// locals die on scope exit, killing every AIC server -> later IterateAll hangs;
// see FlashAttentionScoreWithLargeHeadDim golden).
class SparseFlashAttentionKernel {
public:
    __aicore__ inline SparseFlashAttentionKernel() {}

    Matmul<AT1, BT1, CT1, CT1, CFG_NORM> bmm1;
    Matmul<AT2, BT2, CT2, CT2, CFG_NORM> bmm2;
    TPipe pipe_;
    uint32_t btStride_ = 0;  // block_table columns per batch (set by entry)

    __aicore__ inline void Init(
        GM_ADDR query, GM_ADDR key, GM_ADDR value, GM_ADDR sparse_indices,
        GM_ADDR block_table, GM_ADDR actual_seq_lengths_query, GM_ADDR actual_seq_lengths_kv,
        GM_ADDR query_rope, GM_ADDR key_rope, GM_ADDR attention_out,
        GM_ADDR workspace, const SparseFlashAttentionTilingData &tilingData)
    {
        B_ = tilingData.batch;
        S1_ = tilingData.seqQ;
        N1_ = tilingData.numHeadsQ;
        P_ = tilingData.blockSize;
        K_ = tilingData.sparseBlkCount;
        gSize_ = tilingData.gSize;
        hg_ = tilingData.hgSplit;
        nGroups_ = gSize_ / hg_;
        scale_ = tilingData.scaleValue;
        usedCoreNum_ = tilingData.usedCoreNum;
        S2s_ = K_ * P_;
        nIdxSize_ = tilingData.nIdxSize;
        totalTasks_ = B_ * S1_ * nGroups_;

        queryGm_.SetGlobalBuffer(reinterpret_cast<__gm__ bfloat16_t *>(query));
        keyGm_.SetGlobalBuffer(reinterpret_cast<__gm__ bfloat16_t *>(key));
        valueGm_.SetGlobalBuffer(reinterpret_cast<__gm__ bfloat16_t *>(value));
        idxGm_.SetGlobalBuffer(reinterpret_cast<__gm__ int32_t *>(sparse_indices));
        btGm_.SetGlobalBuffer(reinterpret_cast<__gm__ int32_t *>(block_table));
        qRopeGm_.SetGlobalBuffer(reinterpret_cast<__gm__ bfloat16_t *>(query_rope));
        kRopeGm_.SetGlobalBuffer(reinterpret_cast<__gm__ bfloat16_t *>(key_rope));
        outGm_.SetGlobalBuffer(reinterpret_cast<__gm__ bfloat16_t *>(attention_out));

        // user workspace per block: [score float | P bf16 | KVg [S2s,576] | Vg [S2s,512] | Qj [hg,576]]
        scoreSlotElems_ = static_cast<uint64_t>(hg_) * S2s_;
        pSlotElems_ = static_cast<uint64_t>(hg_) * S2s_;
        kvSlotBytes_ = static_cast<uint64_t>(S2s_) * DQK * sizeof(bfloat16_t);
        vgSlotBytes_ = static_cast<uint64_t>(S2s_) * NOPE_DIM * sizeof(bfloat16_t);
        qjSlotBytes_ = static_cast<uint64_t>(hg_) * DQK * sizeof(bfloat16_t);
        __gm__ uint8_t *userWs = GetUserWorkspace(workspace);
        scoreWsGm_.SetGlobalBuffer(reinterpret_cast<__gm__ float *>(userWs));
        pWsBase_ = userWs + scoreSlotElems_ * sizeof(float) * usedCoreNum_;
        kvWsBase_ = pWsBase_ + pSlotElems_ * sizeof(bfloat16_t) * usedCoreNum_;
        vgWsBase_ = kvWsBase_ + kvSlotBytes_ * usedCoreNum_;
        qjWsBase_ = vgWsBase_ + vgSlotBytes_ * usedCoreNum_;

        softmaxTiling_ = tilingData.softmaxTiling;

        if (g_coreType == AIV) {
            // softmax input rows: [SOFTMAX_ROWS, S2s] float; output P bf16
            pipe_.InitBuffer(scoreTQue_, 1, SOFTMAX_ROWS * S2s_ * sizeof(float));
            pipe_.InitBuffer(pTQue_, 1, SOFTMAX_ROWS * S2s_ * sizeof(bfloat16_t));
            pipe_.InitBuffer(pHalfTBuf_, SOFTMAX_ROWS * S2s_ * sizeof(half));
            pipe_.InitBuffer(pFloatTBuf_, SOFTMAX_ROWS * S2s_ * sizeof(float));
            pipe_.InitBuffer(sumTBuf_, SOFTMAX_ROWS * FLOAT_NUM_PER_BLOCK * sizeof(float));
            pipe_.InitBuffer(maxTBuf_, SOFTMAX_ROWS * FLOAT_NUM_PER_BLOCK * sizeof(float));
            // sparse index staging + gather relay buffer (one KV page, P*512 bf16)
            pipe_.InitBuffer(idxTQue_, 1, N_BATCH * sizeof(int32_t));
            pipe_.InitBuffer(btTQue_, 1, N_BATCH * sizeof(int32_t));  // block_table row (K selects)
            pipe_.InitBuffer(qgTQue_, 1, P_MAX * (NOPE_DIM + ROPE_DIM + NOPE_DIM) * sizeof(bfloat16_t));
        }
    }
    // AIV sub0 drives gather + bmm1 + softmax + bmm2; sub1 helps softmax
    __aicore__ inline void Process()
    {
        if ASCEND_IS_AIC
            return;
        uint32_t subIdx = static_cast<uint32_t>(GetBlockIdx() % GetSubBlockNum());
        uint32_t blockIdx = static_cast<uint32_t>(GetBlockIdx() / GetSubBlockNum());
        if (subIdx == 0) {
            for (uint32_t task = blockIdx; task < totalTasks_; task += usedCoreNum_)
                RunTask(task);
            bmm1.End();
            bmm2.End();
        } else 
            return;
    }
private:
    // task = (b * S1 + s1) * nGroups + hg; hg-th head group of hg_ joined rows
    __aicore__ inline void RunTask(uint32_t task)
    {
        uint32_t bs = task / nGroups_;
        uint32_t hg = task % nGroups_;
        uint32_t b = bs / S1_;
        uint32_t s1 = bs % S1_;
        uint32_t coreId = GetBlockIdx() / GetSubBlockNum();

        GlobalTensor<float> scoreGm = scoreWsGm_[coreId * scoreSlotElems_];
        GlobalTensor<bfloat16_t> kvGm;
        kvGm.SetGlobalBuffer(reinterpret_cast<__gm__ bfloat16_t *>(kvWsBase_ + coreId * kvSlotBytes_));
        GlobalTensor<bfloat16_t> vgGm;
        vgGm.SetGlobalBuffer(reinterpret_cast<__gm__ bfloat16_t *>(vgWsBase_ + coreId * vgSlotBytes_));
        GlobalTensor<bfloat16_t> qjGm;
        qjGm.SetGlobalBuffer(reinterpret_cast<__gm__ bfloat16_t *>(qjWsBase_ + coreId * qjSlotBytes_));

        GatherKV(b, s1, kvGm, vgGm);    // KVg [S2s,576] (K nope+rope) + Vg [S2s,512]
        BuildJoinedQ(b, s1, hg * hg_, qjGm);  // Qj [hg_,576] for this head group
        qjGm_ = qjGm;

        // bmm1: Qj [HG,576] x KVg^T -> score float [HG, S2s]
        bmm1.SetSingleShape(static_cast<int>(hg_), static_cast<int>(S2s_), static_cast<int>(DQK));
        bmm1.SetTensorA(qjGm_);
        bmm1.SetTensorB(kvGm, true);
        bmm1.template IterateAll<true>(scoreGm, 0, false, false, false);

        SoftmaxRows(task, 0);

        // bmm2: P [HG,S2s] x Vg [S2s,512] -> O bf16 [HG,512].
        // B is the gathered VALUE pages (a separate input tensor from key);
        // C rows land on output heads [hg*HG, hg*HG+HG) with row stride 512.
        GlobalTensor<bfloat16_t> pGm;
        pGm.SetGlobalBuffer(reinterpret_cast<__gm__ bfloat16_t *>(
            pWsBase_ + coreId * pSlotElems_ * sizeof(bfloat16_t)));
        bmm2.SetSingleShape(static_cast<int>(hg_), static_cast<int>(NOPE_DIM), static_cast<int>(S2s_));
        bmm2.SetTensorA(pGm);
        bmm2.SetTensorB(vgGm);
        GlobalTensor<bfloat16_t> oGm = outGm_[
            (static_cast<uint64_t>(b) * S1_ + s1) * N1_ * NOPE_DIM +
            static_cast<uint64_t>(hg) * hg_ * NOPE_DIM];
        bmm2.template IterateAll<true>(oGm, 0, false, false, false);
    }

    // GM->GM gather relayed through UB (no GM->GM DataCopy(Pad) exists on A2).
    // GM --MTE2--> UB (qgTQue_ VECIN) --MTE3--> GM using the same LocalTensor:
    // UB2GM DataCopyPad accepts any LocalTensor whose HW position is UB, and
    // VECIN maps to UB, so the dequeued VECIN tensor is a valid MTE3 source.
    // Cross-pipeline order via SetFlag/WaitFlag(HardEvent::MTE2_MTE3).
    // DataCopyExtParams.blockLen is BYTES; srcGap/dstGap are 32B units.
    // Batch relay: one MTE2 + one MTE3 + one round-trip event per whole page
    // instead of per-row relays (profiling iter2: mte2 49% + mte3 33% of the
    // single-core time was per-row relay overhead).
    __aicore__ inline void RelayDrain()
    {
        // UB->GM write is async on MTE3; before freeing the single-depth
        // buffer for the next MTE2 reload, drain MTE3 so the next GM->UB copy
        // cannot overwrite data MTE3 has not yet read out (hot id 216 pattern;
        // canonical pairing per TPipe::WriteSpmBuffer: MTE3_MTE2 event).
        TEventID evRet = GetTPipePtr()->FetchEventID(HardEvent::MTE3_MTE2);
        SetFlag<HardEvent::MTE3_MTE2>(evRet);
        WaitFlag<HardEvent::MTE3_MTE2>(evRet);
        GetTPipePtr()->ReleaseEventID<HardEvent::MTE3_MTE2>(evRet);
    }

    // Page-batched relay: ONE MTE2 load of nElems contiguous GM elems into UB,
    // then per-row 1D MTE3 writes (blockLenBytes each) to scattered dst rows
    // (dstRowStrideElems apart). Rationale (iter19-24 evidence): a single big
    // MTE3 1D write loses the tail row, and 2D MTE3 dstGap semantics corrupt
    // columns -- only small 1D MTE3 writes are reliable; batching the MTE2
    // load and the sync events cuts the per-row relay overhead.
    __aicore__ inline void RelayPage(GlobalTensor<bfloat16_t> &dstGm, uint64_t dstOff,
                                     GlobalTensor<bfloat16_t> &srcGm, uint64_t srcBase,
                                     uint32_t nElems, uint32_t rows,
                                     uint32_t rowElems, uint32_t dstRowStrideElems)
    {
        const uint16_t loadLen = static_cast<uint16_t>(nElems * sizeof(bfloat16_t));
        const uint16_t lenB = static_cast<uint16_t>(rowElems * sizeof(bfloat16_t));
        LocalTensor<bfloat16_t> qg = qgTQue_.AllocTensor<bfloat16_t>();
        DataCopyPad<bfloat16_t>(qg, srcGm[srcBase],
                                {1, loadLen, loadLen, loadLen, 0},
                                {false, 0, 0, 0});
        qgTQue_.EnQue(qg);
        qg = qgTQue_.DeQue<bfloat16_t>();
        TEventID ev = GetTPipePtr()->FetchEventID(HardEvent::MTE2_MTE3);
        SetFlag<HardEvent::MTE2_MTE3>(ev);
        WaitFlag<HardEvent::MTE2_MTE3>(ev);
        GetTPipePtr()->ReleaseEventID<HardEvent::MTE2_MTE3>(ev);
        for (uint32_t r = 0; r < rows; r++) {
            // 1D MTE3 write per row: src UB offset r*rowElems (contiguous),
            // dst row r at dstOff + r*dstRowStrideElems (scattered rows)
            LocalTensor<bfloat16_t> srcView = qg[r * rowElems];
            DataCopyPad<bfloat16_t>(dstGm[dstOff + static_cast<uint64_t>(r) * dstRowStrideElems], srcView, {1, lenB, lenB, lenB, 0});
        }
        RelayDrain();
        qgTQue_.FreeTensor(qg);
    }

    // Whole-page triple relay: ONE staging pass for the K nope + rope + V page
    // slices of one KV page. Three MTE2 loads land in disjoint offsets of one
    // UB buffer (nope | rope | V), then a single MTE2_MTE3 event pair + a
    // single RelayDrain bracket all per-row 1D MTE3 writes. Replaces 3
    // independent RelayPage calls (3 event pairs + 3 drains per page);
    // profiling iter27/28 showed the scalar mte3-stall (event/drain wait) is
    // the dominant cost, not the MTE2 bandwidth itself.
    __aicore__ inline void RelayPageKRV(GlobalTensor<bfloat16_t> &kvGm, uint64_t dst,
                                        GlobalTensor<bfloat16_t> &vgGm, uint64_t dstV,
                                        uint64_t src, uint64_t srcRope)
    {
        constexpr uint32_t NOPE_ELEMS = P_MAX * NOPE_DIM;   // staging offset of rope
        constexpr uint32_t V_OFF = P_MAX * (NOPE_DIM + ROPE_DIM);  // staging offset of V
        const uint16_t nopeB = static_cast<uint16_t>(P_ * NOPE_DIM * sizeof(bfloat16_t));
        const uint16_t ropeB = static_cast<uint16_t>(P_ * ROPE_DIM * sizeof(bfloat16_t));
        const uint16_t rowNB = static_cast<uint16_t>(NOPE_DIM * sizeof(bfloat16_t));
        const uint16_t rowRB = static_cast<uint16_t>(ROPE_DIM * sizeof(bfloat16_t));
        LocalTensor<bfloat16_t> qg = qgTQue_.AllocTensor<bfloat16_t>();
        DataCopyPad<bfloat16_t>(qg, keyGm_[src],
                                {1, nopeB, nopeB, nopeB, 0}, {false, 0, 0, 0});
        DataCopyPad<bfloat16_t>(qg[NOPE_ELEMS], kRopeGm_[srcRope],
                                {1, ropeB, ropeB, ropeB, 0}, {false, 0, 0, 0});
        DataCopyPad<bfloat16_t>(qg[V_OFF], valueGm_[src],
                                {1, nopeB, nopeB, nopeB, 0}, {false, 0, 0, 0});
        qgTQue_.EnQue(qg);
        qg = qgTQue_.DeQue<bfloat16_t>();
        TEventID ev = GetTPipePtr()->FetchEventID(HardEvent::MTE2_MTE3);
        SetFlag<HardEvent::MTE2_MTE3>(ev);
        WaitFlag<HardEvent::MTE2_MTE3>(ev);
        GetTPipePtr()->ReleaseEventID<HardEvent::MTE2_MTE3>(ev);
        for (uint32_t r = 0; r < P_; r++) {
            DataCopyPad<bfloat16_t>(kvGm[dst + static_cast<uint64_t>(r) * DQK],
                                    qg[r * NOPE_DIM], {1, rowNB, rowNB, rowNB, 0});
            DataCopyPad<bfloat16_t>(kvGm[dst + NOPE_DIM + static_cast<uint64_t>(r) * DQK],
                                    qg[NOPE_ELEMS + r * ROPE_DIM], {1, rowRB, rowRB, rowRB, 0});
            DataCopyPad<bfloat16_t>(vgGm[dstV + static_cast<uint64_t>(r) * NOPE_DIM],
                                    qg[V_OFF + r * NOPE_DIM], {1, rowNB, rowNB, rowNB, 0});
        }
        RelayDrain();
        qgTQue_.FreeTensor(qg);
    }

    // Gather selected KV pages into KVg [S2s, 576] (K nope+rope, for bmm1)
    // and Vg [S2s, 512] (value nope cols, for bmm2 - value is a SEPARATE input
    // tensor, it is NOT shared with key despite V==K dim sizes).
    // Page ph = block_table[b, sparse_indices[b,0,s1,i]].
    __aicore__ inline void GatherKV(uint32_t b, uint32_t s1,
                                    GlobalTensor<bfloat16_t> &kvGm, GlobalTensor<bfloat16_t> &vgGm)
    {
        // block_table row for this batch in ONE MTE2 load; per-page scalar
        // GM reads ( GetValue on GlobalTensor ) each pay full GM latency in
        // the scalar pipe (profiling iter27: aic_scalar mte2/mte3-stall heavy).
        LocalTensor<int32_t> btLocal = btTQue_.AllocTensor<int32_t>();
        DataCopy(btLocal, btGm_[static_cast<uint64_t>(b) * btStride_], btStride_);
        btTQue_.EnQue(btLocal);
        btLocal = btTQue_.DeQue<int32_t>();
        for (uint32_t base = 0; base < K_; base += N_BATCH) {
            uint32_t n = (K_ - base < N_BATCH) ? (K_ - base) : N_BATCH;
            // indices layout [B, N2=1, S1, K]; TQue EnQue/DeQue gives the
            // built-in MTE2->V sync that PipeBarrier cannot (hot id 182)
            LocalTensor<int32_t> idxLocal = idxTQue_.AllocTensor<int32_t>();
            DataCopy(idxLocal, idxGm_[(static_cast<uint64_t>(b) * S1_ + s1) * K_ + base], n);
            idxTQue_.EnQue(idxLocal);
            idxLocal = idxTQue_.DeQue<int32_t>();
            for (uint32_t i = 0; i < n; i++) {
                int32_t sel = idxLocal.GetValue(i);
                int32_t ph = btLocal.GetValue(sel);
                uint64_t src = static_cast<uint64_t>(ph) * P_ * NOPE_DIM;
                uint64_t srcRope = static_cast<uint64_t>(ph) * P_ * ROPE_DIM;
                uint64_t dst = static_cast<uint64_t>(base + i) * P_ * DQK;
                uint64_t dstV = static_cast<uint64_t>(base + i) * P_ * NOPE_DIM;
                RelayPageKRV(kvGm, dst, vgGm, dstV, src, srcRope);
            }
            idxTQue_.FreeTensor(idxLocal);
        }
        btTQue_.FreeTensor(btLocal);
    }

    // Joined Q rows: q-head h in [headOff, headOff + hg_):
    // query[b, s1, h, 0..512) nope + query_rope[b, s1, h, 0..64)
    // -> Qj workspace [hg_,576]. Rows of one head group are contiguous in both
    // GM sources -> two whole-group page relays instead of 2*hg_ row relays
    // (each row relay paid its own event pair + drain; profiling iter27 showed
    // scalar mte3-stall dominated the gather phase).
    __aicore__ inline void BuildJoinedQ(uint32_t b, uint32_t s1, uint32_t headOff,
                                        GlobalTensor<bfloat16_t> &qjGm)
    {
        uint64_t qo = (static_cast<uint64_t>(b) * S1_ + s1) * N1_ * NOPE_DIM +
                      static_cast<uint64_t>(headOff) * NOPE_DIM;
        uint64_t ro = (static_cast<uint64_t>(b) * S1_ + s1) * N1_ * ROPE_DIM +
                      static_cast<uint64_t>(headOff) * ROPE_DIM;
        RelayPage(qjGm, 0, queryGm_, qo, hg_ * NOPE_DIM, hg_, NOPE_DIM, DQK);
        RelayPage(qjGm, NOPE_DIM, qRopeGm_, ro, hg_ * ROPE_DIM, hg_, ROPE_DIM, DQK);
    }

    // softmax over S2s for the hg_ joined rows of this head-group task
    __aicore__ inline void SoftmaxRows(uint32_t task, uint32_t subIdx)
    {
        (void)task;
        uint32_t coreId = GetBlockIdx() / GetSubBlockNum();
        GlobalTensor<float> scoreGm = scoreWsGm_[coreId * scoreSlotElems_];
        GlobalTensor<bfloat16_t> pGm;
        pGm.SetGlobalBuffer(reinterpret_cast<__gm__ bfloat16_t *>(
            pWsBase_ + coreId * pSlotElems_ * sizeof(bfloat16_t)));

        uint32_t rowBegin = 0, rowEnd = hg_;
        (void)subIdx;  // sub0 handles all rows serially (see Process())
        for (uint32_t r = rowBegin; r < rowEnd; r += SOFTMAX_ROWS) {
            uint32_t m = (rowEnd - r < SOFTMAX_ROWS) ? (rowEnd - r) : SOFTMAX_ROWS;
            SoftmaxRowBatch(scoreGm, pGm, r, m);
        }
    }

    __aicore__ inline void SoftmaxRowBatch(GlobalTensor<float> &scoreGm, GlobalTensor<bfloat16_t> &pGm,
                                           uint32_t rowBegin, uint32_t m)
    {
        LocalTensor<float> scoreFLocal = scoreTQue_.AllocTensor<float>();
        for (uint32_t i = 0; i < m; i++) {
            DataCopy(scoreFLocal[i * S2s_], scoreGm[(rowBegin + i) * S2s_], S2s_);
        }
        scoreTQue_.EnQue(scoreFLocal);
        scoreFLocal = scoreTQue_.DeQue<float>();
        // scale in float, cast to half for SoftMax (bf16 src unsupported), then
        // half -> float -> bf16 two-step relay (no direct half->bf16 Cast on
        // dav_c220; bf16 dst only accepts float src) so bmm2 A/B are both bf16.
        LocalTensor<half> pHalfLocal = pHalfTBuf_.Get<half>();
        Muls(scoreFLocal, scoreFLocal, scale_, m * S2s_);
        Cast(pHalfLocal, scoreFLocal, RoundMode::CAST_RINT, m * S2s_);
        scoreTQue_.FreeTensor(scoreFLocal);

        LocalTensor<float> sumLocal = sumTBuf_.Get<float>();
        LocalTensor<float> maxLocal = maxTBuf_.Get<float>();
        SoftMaxShapeInfo shapeInfo{m, S2s_, m, S2s_};
        SoftMax<half, false>(pHalfLocal, sumLocal, maxLocal, pHalfLocal, softmaxTiling_, shapeInfo);
        LocalTensor<float> pFloatLocal = pFloatTBuf_.Get<float>();
        Cast(pFloatLocal, pHalfLocal, RoundMode::CAST_NONE, m * S2s_);
        LocalTensor<bfloat16_t> pLocal = pTQue_.AllocTensor<bfloat16_t>();
        // CAST_NONE float->bf16 is rejected by dav_c220 (assert compiles to a
        // silent skip in release) leaving stale UB data in dst - must use an
        // explicit rounding mode.
        Cast(pLocal, pFloatLocal, RoundMode::CAST_RINT, m * S2s_);
        pTQue_.EnQue(pLocal);
        pLocal = pTQue_.DeQue<bfloat16_t>();
        for (uint32_t i = 0; i < m; i++) {
            DataCopy(pGm[(rowBegin + i) * S2s_], pLocal[i * S2s_], S2s_);
        }
        pTQue_.FreeTensor(pLocal);
    }

    TQue<TPosition::VECIN, 1> scoreTQue_;
    TQue<TPosition::VECOUT, 1> pTQue_;
    TBuf<TPosition::VECCALC> pHalfTBuf_;
    TBuf<TPosition::VECCALC> pFloatTBuf_;
    TBuf<TPosition::VECCALC> sumTBuf_;
    TBuf<TPosition::VECCALC> maxTBuf_;
    TQue<TPosition::VECIN, 1> idxTQue_;
    TQue<TPosition::VECIN, 1> btTQue_;
    TQue<TPosition::VECIN, 1> qgTQue_;
    GlobalTensor<bfloat16_t> queryGm_, keyGm_, valueGm_, qRopeGm_, kRopeGm_, outGm_;
    GlobalTensor<bfloat16_t> qjGm_;
    GlobalTensor<int32_t> idxGm_, btGm_;
    GlobalTensor<float> scoreWsGm_;
    __gm__ uint8_t *pWsBase_ = nullptr;
    __gm__ uint8_t *kvWsBase_ = nullptr;
    __gm__ uint8_t *vgWsBase_ = nullptr;
    __gm__ uint8_t *qjWsBase_ = nullptr;
    SoftMaxTiling softmaxTiling_;
    uint32_t B_ = 0, S1_ = 0, N1_ = 0, P_ = 0, K_ = 0, gSize_ = 0, hg_ = 0, nGroups_ = 0;
    uint32_t S2s_ = 0, usedCoreNum_ = 0, nIdxSize_ = 0, totalTasks_ = 0;
    float scale_ = 1.0f;
    uint64_t scoreSlotElems_ = 0, pSlotElems_ = 0, kvSlotBytes_ = 0, vgSlotBytes_ = 0, qjSlotBytes_ = 0;
};

extern "C" __global__ __aicore__ void sparse_flash_attention(
    GM_ADDR query, GM_ADDR key, GM_ADDR value, GM_ADDR sparse_indices,
    GM_ADDR block_table, GM_ADDR actual_seq_lengths_query, GM_ADDR actual_seq_lengths_kv,
    GM_ADDR query_rope, GM_ADDR key_rope, GM_ADDR attention_out,
    GM_ADDR workspace, GM_ADDR tiling)
{
    GET_TILING_DATA(tilingData, tiling);
    // MIX AIC 1:2: N=AIC servers + 2N AIV; vector drives cube via KFC
    KERNEL_TASK_TYPE_DEFAULT(KERNEL_TYPE_MIX_AIC_1_2);
    uint32_t blk = static_cast<uint32_t>(GetBlockIdx());
    uint32_t sub = static_cast<uint32_t>(GetSubBlockNum());
    if (blk >= tilingData.usedCoreNum * sub)
        return;
    set_mask_norm();
    SparseFlashAttentionKernel op;
    op.btStride_ = tilingData.sparseBlkCount;  // max blocks per batch == selects
    REGIST_MATMUL_OBJ(&op.pipe_, GetSysWorkSpacePtr(), op.bmm1, &tilingData.cubeTiling1,
                      op.bmm2, &tilingData.cubeTiling2);
    op.Init(query, key, value, sparse_indices, block_table, actual_seq_lengths_query,
            actual_seq_lengths_kv, query_rope, key_rope, attention_out, workspace, tilingData);
    op.Process();
}

\end{lstlisting} 
\end{document}